%% file: iclr2027_conference_v_Michele.tex
\documentclass{article} % For LaTeX2e
\usepackage{iclr2027_conference,times}

\input{math_commands.tex}

\usepackage[T1]{fontenc}
\usepackage[utf8]{inputenc} % harmless with pdfLaTeX; optional in newer LaTeX
\usepackage{microtype}

\usepackage{amsmath}
\usepackage{amssymb}
\usepackage{amsfonts}
\usepackage{mathtools}

\usepackage{booktabs}
\usepackage{tabularx}
\usepackage{array}
\usepackage{longtable}
\usepackage{makecell}
\usepackage[table]{xcolor}

\usepackage{graphicx}
\usepackage{subcaption}
\usepackage{pgfplots}
\pgfplotsset{compat=1.18}

\usepackage{enumitem}
\usepackage{pifont}

\usepackage{listings}

\definecolor{codegray}{RGB}{245,245,245}
\definecolor{codeblue}{RGB}{30,70,130}

\usepackage{url}
\usepackage{hyperref}
\usepackage[nameinlink,noabbrev]{cleveref}

\newcommand{\cmark}{\ding{51}} % yes
\newcommand{\xmark}{\ding{55}} % no
\newcommand{\pmark}{\textcolor{gray}{\(\triangle\)}} % partial / indirect

\newcommand{\MoSAR}{\textsc{MoSAR}}
\newcommand{\DeltaSimplex}{\Delta^{R-1}}

\newcommand{\GELU}{\operatorname{GELU}}
\newcommand{\RoPE}{\operatorname{RoPE}}
\newcommand{\vecop}{\operatorname{vec}}

\newcommand{\code}[1]{\texttt{#1}}
\newcommand{\codepath}[1]{\path{#1}}
\DeclareMathOperator{\Att}{Att}
\DeclareMathOperator{\SparseAtt}{SparseAtt}

\title{MoSAR: Mixture of Semantic Attention Regimes for Learning Adaptive and Approximable Attention Geometries}

\author{
Michele Paolicelli \\
Department of Computer Science \\
Università degli Studi di Bari Aldo Moro \\
Bari, 70121, Italy \\
\texttt{m.paolicelli9@phd.uniba.it} \And
Alessandro Petruzzelli \\
Department of Computer Science \\
Università degli Studi di Bari Aldo Moro \\
Bari, 70121, Italy \\
\texttt{alessandro.petruzzelli@uniba.it} \And
Alessandro Franceso Maria Martina \\
Department of Computer Science \\
Università degli Studi di Bari Aldo Moro \\
Bari, 70121, Italy \\
\texttt{alessandro.martina@uniba.it} \And
Cataldo Musto \\
Department of Computer Science \\
Università degli Studi di Bari Aldo Moro \\
Bari, 70121, Italy \\
\texttt{cataldo.musto@uniba.it} \And
Giovanni Semeraro \\
Department of Computer Science \\
Università degli Studi di Bari Aldo Moro \\
Bari, 70121, Italy \\
\texttt{giovanni.semeraro@uniba.it} \And
}

\iclrfinalcopy % Uncomment for camera-ready version, but NOT for submission.
\begin{document}

\maketitle

\begin{abstract}

The quadratic complexity of dense self-attention remains a central bottleneck for long-context language modeling. Many efficient alternatives address this cost by deciding in advance where attention should be sparse or local. We argue that attention approximation should instead be approached as a geometric problem, with the relevant interaction geometry learned from data: natural-language dependencies are input-dependent and difficult to prescribe in advance, so the model should learn where positional relevance can decay and where broader interactions must be preserved. We introduce Mixture of Semantic Attention Regimes (MoSAR), which learns such an adaptive, controlled-decay geometry over query--key interactions. Input-conditioned query and key routers, applied after positional encoding, select mixtures over short, medium, and global regimes, inducing a continuous distance-dependent attention field rather than a fixed sparsity pattern. This geometry is learned during training and can subsequently be discretized through top-1 routing. In controlled pre-training experiments with matched 500M-parameter models, MoSAR learns a substantially lower-reach attention geometry without degrading language-modeling quality, improving perplexity over dense RoPE at the training context length. Under length extrapolation, MoSAR achieves the best perplexity among all evaluated variants, including strong baselines such as ALiBi. Moreover, the learned geometry remains stable under deterministic top-1 discretization, suggesting that it is not only adaptive, but also amenable to low-cost approximation at inference time.

\end{abstract}

\section{Introduction}

\begin{figure}[t]
    \centering
    \includegraphics[width=0.65\linewidth]{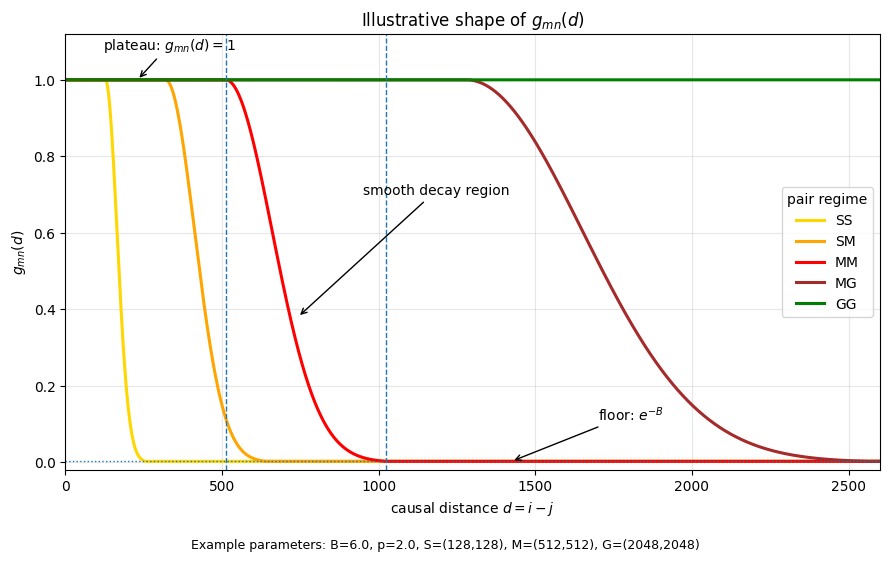}
    \caption{
    Illustrative MoSAR decay profiles for representative query--key regime pairs.
Each pair induces a distance-dependent regime $g_{mn}(d)$ with an initial
unit plateau, a smooth decay region, and a finite floor; pairs
involving broader regimes retain interactions over longer causal distances.
    }
    \label{fig:mosar-decay-kernel}
\end{figure}

Self-attention~\citep{vaswani2017attention} is a central component of modern Large Language Models (LLMs)~\citep{minaee2024large}, but its quadratic complexity in sequence length remains a major obstacle to efficient long-context modeling. A large body of work has therefore explored a plethora of alternatives to dense attention~\citep{tay2022efficient}. Many of these approaches reduce computation by prescribing a fixed sparsity pattern, such as a sliding window or a predetermined set of admissible interactions~\citep{beltagy2020longformer,zaheer2020big,chaplot2023albert}. While effective, such strategies make a strong assumption: the same notion of locality is appropriate for every token, layer, and context. In practice, dense attention remains widely used in modern LLMs, supported by increasingly optimized implementations~\citep{dao2022flashattention,dao2023flashattention,kwon2023efficient}.

On the other hand, standard positional mechanisms determine how token interactions depend on relative position and therefore induce a geometry over attention. Rotary Positional Encoding (RoPE)~\citep{su2021roformer}, for instance, provides an effective relative positional geometry through rotations of queries and keys, but its positional interaction is oscillatory rather than explicitly decaying with distance~\citep{barbero2024round}. This is not a limitation of RoPE itself---decay is not what RoPE was designed to provide---but it raises a natural question: \emph{how, and when, should positional relevance decay?} A single decay law would still impose one universal attention geometry. Language dependencies occur at different scales, and there is no reason to assume that the same locality regime should be optimal for every token or at every layer.
Therefore, we ask a different question: \emph{can the model learn from data where attention computation should remain local and where broader interactions are worth preserving?}

In this work, we take a different perspective and argue that attention approximation is fundamentally a \emph{geometric} problem: sparsity should not necessarily be imposed \emph{a priori}; it can instead emerge from a distance-dependent decay geometry learned by the model. We therefore introduce \textbf{Mixture of Semantic Attention Regimes (MoSAR)}, an adaptive attention mechanism in which post-RoPE query--key routers select among multiple smoothly decaying functions $g_{mn}(d)$ called \textit{semantic regimes} (Figure \ref{fig:mosar-decay-kernel}). Each regime defines three aspects of the resulting geometry: a locality region in which RoPE-based interactions are largely preserved, a transition region in which their contribution progressively decays, and a low-relevance region that can eventually be treated as computationally negligible. Thus, MoSAR replaces a single fixed sparse pattern with a family of approximable geometries selected dynamically by the model. We also study \textbf{MoSAR-Cost}, a cost-aware variant that explicitly biases the routers toward shorter expected reach. Throughout the paper, we use \textit{semantic} to emphasize that regime selection is conditioned on learned contextual representations, rather than solely on position or a fixed sparsity pattern. Introducing distance-dependent decay can provide at least three desirable properties. First, it augments relative positional information with an inductive bias on relative relevance. Second, a smooth decay avoids the abrupt \textit{semantic discontinuity} introduced by hard local masks, acting as a geometric regularizer. Third, sufficiently decayed interactions expose regions of the attention matrix that can be approximated or omitted altogether.

Our main contributions are:

\begin{enumerate}
\item We formulate efficient attention as a \textbf{controlled-decay geometry problem}, and introduce \textbf{Mixture of Semantic Attention Regimes (MoSAR)}, in which post-RoPE bilinear query and key routers select adaptive decay geometries.

\item We show, through controlled 500M-parameter from-scratch pre-training, that, among the evaluated variants, \textbf{adaptive controlled decay improves over dense RoPE and fixed locality baselines}, with MoSAR achieving the best perplexity at 4096 and 8192 tokens.

\item We demonstrate that the learned soft geometry is \textbf{amenable to approximation at inference time}: deterministic top-1 routing substantially reduces the expected normalized reach while preserving competitive perplexity under length extrapolation; MoSAR-Cost further shows that \textbf{explicit reach pressure can move this trade-off toward substantially lower expected reach}.

\end{enumerate}

\section{Background and Related Work}
% ============================================================
% Gap-analysis table
% ============================================================
\begin{table*}[t]
    \centering
    \caption{
        Comparison of methods that modify attention geometry or reduce
        attention computation.
        Columns refer to the mechanism introduced beyond standard
        content-based self-attention.
        \cmark{} denotes an explicit property,
        \xmark{} its absence, and
        \pmark{} a partial, indirect, or inference-only property.
    }
    \label{tab:mosar-gap-analysis}

    \setlength{\tabcolsep}{3.6pt}
    \renewcommand{\arraystretch}{1.12}
    \scriptsize

    \resizebox{\textwidth}{!}{
    \begin{tabular}{@{}lcccccccc@{}}
        \toprule

        \textbf{Method}
        &
        \makecell{\textbf{Data-}\\\textbf{learned}}
        &
        \makecell{\textbf{Input}\\\textbf{adaptive}}
        &
        \makecell{\textbf{Token-}\\\textbf{wise}}
        &
        \makecell{\textbf{Attention}\\\textbf{approx.}}
        &
        \makecell{\textbf{Cost}\\\textbf{loss}}
        &
        \makecell{\textbf{Learned}\\\textbf{routing}}
        &
        \makecell{\textbf{Distance}\\\textbf{decay}}
        &
        \makecell{\textbf{Learned}\\\textbf{decay}}
        \\

        \midrule

        \multicolumn{9}{@{}l}{
            \textit{Efficient and approximate attention}
        }
        \\[1pt]

        \textsc{Reformer}~\citep{kitaev2020reformer}
        & \pmark & \cmark & \cmark & \cmark
        & \xmark & \xmark & \xmark & \xmark
        \\

        \textsc{Linformer}~\citep{wang2020linformer}
        & \cmark & \xmark & \xmark & \cmark
        & \xmark & \xmark & \xmark & \xmark
        \\

        \textsc{Longformer}~\citep{beltagy2020longformer}
        & \xmark & \xmark & \xmark & \cmark
        & \xmark & \xmark & \xmark & \xmark
        \\

        \midrule

        \multicolumn{9}{@{}l}{
            \textit{Positional and geometric modifications}
        }
        \\[1pt]

        \textsc{ALiBi}~\citep{press2021train}
        & \xmark & \xmark & \xmark & \xmark
        & \xmark & \xmark & \cmark & \xmark
        \\

        \textsc{NoPE}~\citep{kazemnejad2023impact}
        & \pmark & \xmark & \xmark & \xmark
        & \xmark & \xmark & \xmark & \xmark
        \\

        \textsc{XPos/LEX}~\citep{sun2023length}
        & \xmark & \xmark & \xmark & \pmark
        & \xmark & \xmark & \cmark & \xmark
        \\

        \textsc{YaRN}~\citep{peng2024yarn}
        & \xmark & \xmark & \xmark & \xmark
        & \xmark & \xmark & \xmark & \xmark
        \\

        \(p\)\textsc{-RoPE}~\citep{barbero2024round}
        & \xmark & \xmark & \xmark & \xmark
        & \xmark & \xmark & \xmark & \xmark
        \\

        \textsc{CoPE}~\citep{amballa2025cope}
        & \pmark & \xmark & \xmark & \xmark
        & \xmark & \xmark & \xmark & \xmark
        \\

        \textsc{HoPE}~\citep{dai2025hope}
        & \pmark & \xmark & \xmark & \xmark
        & \xmark & \xmark & \cmark & \pmark
        \\

        \midrule

        \multicolumn{9}{@{}l}{
            \textit{Dynamic selection and sparsification}
        }
        \\[1pt]

        \textsc{Selective Attention}~\citep{leviathan2025selective}
        & \pmark & \cmark & \cmark & \pmark
        & \xmark & \xmark & \xmark & \xmark
        \\

        \textsc{FlexPrefill}~\citep{lai2025flexprefill}
        & \xmark & \cmark & \pmark & \cmark
        & \xmark & \xmark & \xmark & \xmark
        \\

        \textsc{Adaptive Attention Span}~\citep{sukhbaatar2019adaptive}
        & \cmark & \pmark & \pmark & \cmark
        & \cmark & \xmark & \cmark & \pmark
        \\

        \textsc{MoBA}~\citep{lu2026moba}
        & \pmark & \cmark & \cmark & \cmark
        & \xmark & \pmark & \xmark & \xmark
        \\

        \textsc{NSA}~\citep{yuan2025native}
        & \cmark & \cmark & \cmark & \cmark
        & \xmark & \pmark & \xmark & \xmark
        \\

        \midrule

        \rowcolor{gray!12}
        \textbf{MoSAR family (ours)}
        & \cmark & \cmark & \cmark & \cmark
        & \cmark & \cmark & \cmark & \cmark
        \\

        \bottomrule
    \end{tabular}
    }

    \vspace{2pt}
    \begin{minipage}{0.98\textwidth}
        \footnotesize
        \textit{Data-learned} indicates that the relevant geometry or
        selection mechanism is learned end-to-end rather than fully
        prescribed.
        \textit{Attention approximation} denotes an explicit reduction
        of evaluated query--key interactions or an equivalent
        low-rank/sparse computation.
        For MoSAR, decay and approximation are jointly induced by
        token-conditioned regime selection and an explicit
        attention-cost objective.
    \end{minipage}
\end{table*}

The starting point of this work is self-attention (SA)~\citep{vaswani2017attention}
endowed with a positional geometry. With RoPE~\citep{su2021roformer}, relative
positional information is injected by rotating queries and keys before computing
their similarity. Given \(Q,K,V\in\mathbb{R}^{n\times d}\), SA with RoPE computes
\begin{equation}\label{eq:self-attention}
\Att(Q,K,V)
=
\softmax\left(
\frac{
\left(q_i^\top R^d_{\Theta,j-i} k_j\right)_{i,j=1}^{n}
}{\sqrt d}
\right)V,
\end{equation}
where \(R^d_{\Theta,i}\) is a block-diagonal rotation matrix with \(d/2\)
two-dimensional orthogonal blocks,
\[
R^d_{\Theta,i}
=
\left(
    \begin{array}{cc}
        \cos(i\theta_h) & -\sin(i\theta_h) \\
        \sin(i\theta_h) &  \cos(i\theta_h)
    \end{array}
\right)_{1\leq h\leq d/2},
\qquad
\theta_h=10000^{-2(h-1)/d}.
\]
Thus, SA is not only a computation over token similarities: through the
positional mechanism, it induces a geometry over query--key interactions. A common intuition is that useful positional mechanisms should make distant
tokens progressively less relevant. In general, RoPE induce an oscillatory decay of query--key similarity with relative distance, which is not a monotone decay; indeed,
\citet{barbero2024round} argue that such decay is unlikely to be the main reason
for RoPE's effectiveness. ALiBi~\citep{press2021train} provides a complementary
perspective by explicitly biasing query--key attention scores with a penalty
proportional to their distance. XPos~\citep{sun2023length} adds an exponential envelope to
stabilise long-range behaviour. YaRN~\citep{peng2024yarn} rescales RoPE frequencies to extend
a trained context. \(p\)-RoPE~\citep{barbero2024round} truncates low frequencies. Each
prescribes a different law, and each is fixed before training begins. A few methods begin to move the choice into the model. CoPE~\citep{amballa2025cope} makes
position itself contextual, counting gated events rather than tokens, so what counts as
distance depends on content. HoPE~\citep{dai2025hope} mixes retained and decayed frequency
components. Both learn part of the geometry. But they change what distance means, or which
frequencies survive. Neither lets a token choose how fast its own relevance should fall away.

The SA computation in Eq.~\ref{eq:self-attention} has complexity $O(n^2d)$, and a parallel line of work addresses this cost through selection and sparsification.
Longformer~\citep{beltagy2020longformer} and BigBird~\citep{zaheer2020big} fix a pattern in
advance; Reformer~\citep{kitaev2020reformer} buckets by hash; Linformer~\citep{wang2020linformer}
projects keys to low rank. Selective Attention~\citep{leviathan2025selective},
FlexPrefill~\citep{lai2025flexprefill}, MoBA~\citep{lu2026moba} and NSA~\citep{yuan2025native}
adapt the attention pattern dynamically to the input. Their focus, however, is primarily on selecting which query--key interactions should be evaluated, rather than on learning how interaction strength should vary continuously with distance within the selected domain. In several cases, this selection still relies on an explicit or approximate query--key compatibility signal. 

Given the large number of proposed efficient-attention methods, we focus on the
ideas most closely related to the design of MoSAR: positional attention
geometries, sparse and approximate attention, and dynamic mechanisms for
selecting which interactions should remain computationally relevant.
Table~\ref{tab:mosar-gap-analysis} summarizes representative methods across
these directions and is intended as a gap analysis.

These methods motivate the geometric side of
our question: \emph{if positional relevance is not governed by a universal decay law,
how should the support of an efficient attention mechanism be determined?} Positional methods prescribe one profile and apply it everywhere. Selection methods adapt to the input but leave the profile untouched. In particular, Adaptive Attention
Span~\citep{sukhbaatar2019adaptive} gives each head a learnable span, applies a soft ramp at
its edge, and penalises the span in the loss. It is the closest precedent for the claim we
make: the geometry is learned rather than prescribed, and learned under explicit cost
pressure. MoSAR differs in three ways. The span is learned per head and then applies to every
token; MoSAR routes per token, in context. The span is a single scalar; MoSAR selects a
mixture over several regimes. And the span constrains queries only; MoSAR routes queries and
keys separately and couples them, so the profile of a pair depends on both of its ends.

The aforementioned approaches progressively relax fixed
attention patterns, but typically combine only subsets of the properties that
motivate MoSAR. This progression exposes a remaining circularity. To sparsify attention, one
would ideally identify query--key interactions that are unlikely to matter;
however, evaluating their exact compatibility is itself one of the operations
that dense attention performs. This is the catch-22 that motivates MoSAR.
Rather than predicting exact pairwise relevance or imposing a universal sparse
pattern, MoSAR learns token-conditioned query and key routing over a small set
of controlled distance regimes. Appendix~\ref{app:catch22} further formalizes the catch-22 through a conditional sparse-approximation argument and explains how adaptive regimes
relax the assumptions required by fixed-decay constructions.

\section{MoSAR: Mixture of Semantic Attention Regimes}\label{mosar}

A central difficulty in efficient attention is that identifying useful query--key pairs is itself a circular problem. Ideally, one would avoid computing $q_i^\top k_j$ whenever the interaction $(i,j)$ is unlikely to matter; however, deciding whether $(i,j)$ matters by evaluating its exact compatibility already requires forming the dense attention matrix. MoSAR addresses this problem through \emph{bilinear routing}. Instead of routing only queries or only keys, which would respectively select row-wise or column-wise regions, MoSAR assigns locality preferences to both positions $i$ and $j$. Their interaction selects a regime for the pair, allowing the model to localize two-dimensional regions of the attention matrix before the corresponding query--key similarities are computed.

In this sense, MoSAR is related to Mixture-of-Experts conditional computation~\citep{shazeer2017outrageously,jiang2024mixtral,cai2025survey}, but routes \emph{regions of attention computation} rather than tokens to independent experts. Architecturally, MoSAR can be viewed as a lightweight router inserted on the post-RoPE $Q$--$K$ pathway. Conceptually, however, its role is not merely to add another module, but to induce a regime-dependent attention geometry. The router selects among semantic attention regimes, while each regime defines a controlled distance-dependent decay. The resulting geometry determines which interactions should be preserved, which should be attenuated, and which may become computationally negligible. We provide the full implementation-oriented decoder diagram in Appendix~\ref{app:mosar-architecture}; the notation used throughout the section is summarized in Appendix~\ref{notation}.

\paragraph{Semantic Regimes.} A \emph{semantic regime} is a fixed law for how attention falls off with relative distance, whose activation depends on token content. Formally, a semantic regime is a gate function $g_m:\mathbb{N}_0\to[e^{-B},1]$ that has a \emph{plateau} $T_m$, a \emph{semantic transition zone} $L_m$, a \emph{reach}
$\delta_m:=T_m+L_m$ and a \emph{plateau fraction} $\alpha_m$, where $m$ indexes the available regimes. Figure~\ref{fig:mosar-decay-kernel} provides a visual reference for the regime geometry described throughout this paragraph. The semantic regime splits distance into the three zones. Up to
$T_m:=\alpha_m\delta_m$ lies the \emph{locality} zone, where $g_m=1$ and the RoPE-based interaction
passes through unchanged. The next $L_m:=(1-\alpha_m)\delta_m$ positions form the \emph{transition}
zone, where $g_m$ decays smoothly. Beyond $\delta_m$ lies the \emph{low-relevance} zone, where $g_m$
holds at the floor $e^{-B}$. For numerical stability, that floor is finite: distant interactions are attenuated, never
deleted. A regime therefore fixes the scale at which an interaction is evaluated, not which
interaction is relevant; that remains the job of $q_i^\top k_j$. Regimes are fixed before
training. What MoSAR learns is the distribution over them, and so the profile each pair $(q_i,k_j)$
receives. A pair brings two regimes, one from each end. For a query regime $m$ and a key regime $n$ we
average them:
$\delta_{mn}:=\frac{\delta_m+\delta_n}{2},\;
\alpha_{mn}:=\frac{\alpha_m+\alpha_n}{2},\;
T_{mn}:=\alpha_{mn}\delta_{mn},\;
L_{mn}:=(1-\alpha_{mn})\delta_{mn}.$
Averaging gives neither end a veto. For causal distance $d=i-j\ge0$, the averaged regime is:
\[
g_{mn}(d):=
\begin{cases}
1, & d\le T_{mn},\\[3pt]
\exp\left[-B\left(\dfrac{d-T_{mn}}{L_{mn}}\right)^p\right], & T_{mn}<d<T_{mn}+L_{mn},\\[9pt]
e^{-B}, & d\ge T_{mn}+L_{mn}.
\end{cases}
\]
Where $g_{GG}(d)=1$ for all $d\ge0$. We use $R$ regimes, collected in $\mathcal{R}$. By default $\mathcal{R}=\{S,M,G\}$, with
reaches $(\delta_S,\delta_M,\delta_G)=(128,512,2048)$, plateau fractions
$(\alpha_S,\alpha_M,\alpha_G)=(0.75,0.5,1)$, barrier depth $B=6$, and decay exponent $p=2$. A query asking for $S$ paired with a key asking for $G$
lands at reach $1088$: the query is pulled outward, the key inward. Because $\alpha_G=1$, the global regime has
$L_G=0$ and applies no bias at all: $g_G\equiv1$. MoSAR therefore contains dense RoPE
attention as a special case. A token routed to $G$ has chosen not to decay. Collecting the gates gives the distance-conditioned regime
matrix $\mathbf G(d):=\left[g_{mn}(d)\right]_{m,n=1}^{R}$. The default configuration yields six distinct pair
geometries:
\begin{center}\label{geometries}
\begin{tabular}{lrrrrrr}
\toprule
Pair & $SS$ & $SM$ & $MM$ & $SG$ & $MG$ & $GG$ \\
\midrule
$\delta_{mn}$ & 128 & 320 & 512 & 1088 & 1280 & 2048 \\
$T_{mn}$ & 96 & 200 & 256 & 952 & 960 & 2048 \\
$L_{mn}$ & 32 & 120 & 256 & 136 & 320 & 0 \\
\bottomrule
\end{tabular}
\end{center}

\paragraph{Routers MLP.} At layer $\ell$, let $h_i^\ell\in\R^{d_{\mathrm{model}}}$ denote the hidden
state of token $i$. A pre-attention RMS normalization~\citep{zhang2019root} gives
$\bar h_i^\ell = \operatorname{Norm}(h_i^\ell).$ Standard projections produce
$q_i^\ell=W_Q^\ell\bar h_i^\ell,\;k_i^\ell=W_K^\ell\bar h_i^\ell,\;v_i^\ell=W_V^\ell\bar h_i^\ell.$ After reshaping into heads, a backbone-specific positional transformation is
applied: $
\widetilde q_i^{\ell,h}=\Phi_Q^\ell(q_i^{\ell,h},i),\;\widetilde k_i^{\ell,r}=\Phi_K^\ell(k_i^{\ell,r},i).$
By default, $\Phi_Q^\ell=\Phi_K^\ell=\RoPE.$ For query token $i$, and key token $j$, concatenate the post-positional query and key/value heads respectively: $
x_{i,Q}^\ell
=
\vecop\!\left(
\widetilde q_i^{\ell,1},\ldots,
\widetilde q_i^{\ell,H_Q}
\right)
\in\R^{H_Qd_h},\;
x_{j,K}^\ell
=
\vecop\!\left(
\widetilde k_j^{\ell,1},\ldots,
\widetilde k_j^{\ell,H_{KV}}
\right)
\in\R^{H_{KV}d_h}.
$
The key heads must not be repeated to $H_Q$ before the key router. Repetition
is only a transparent reference operation for grouped-query attention after
routing, when computing head-specific similarities. The query router (likewise for the key router) is a small MLP:
\[
\pi_{i,Q}^\ell=\softmax\!\left(W_{2,Q}^\ell\cdot\GELU\!(W_{1,Q}^\ell x_{i,Q}^\ell+b_{1,Q}^\ell)+b_{2,Q}^\ell\right).
\]
Where
$W_{1,Q}^\ell\in\R^{d_r\times H_Qd_h},\;W_{1,K}^\ell\in\R^{d_r\times H_{KV}d_h},\;
W_{2,Q}^\ell,W_{2,K}^\ell\in\R^{R\times d_r}.$ We use GELU activations~\citep{hendrycks2016gaussian}, followed by a softmax over regimes to obtain
$\pi_{i,Q}^\ell\in\DeltaSimplex,$ and $\pi_{j,K}^\ell\in\DeltaSimplex,$ where $\DeltaSimplex$ denotes the probability simplex over the $R$ attention regimes. Componentwise,
$\pi_{i,Q}^\ell(m)\ge 0,$ $\sum_{m=1}^{R}\pi_{i,Q}^\ell(m)=1.$ Likewise for the keys.

\paragraph{MoSAR.}\label{Separable mixture and Head-specific similarities} The \MoSAR{} positive worth field for edge $(i,j)$ is:
\begin{align}\label{worth_field}
W_\theta^\ell(i,j)
&:=\sum_{m=1}^{R}\sum_{n=1}^{R}
\pi_{i,Q}^\ell(m)
\pi_{j,K}^\ell(n)
g_{mn}(i-j)\\
&=\pi_{i,Q}^{\ell\top}\mathbf G(i-j)\pi_{j,K}^\ell.
\end{align}
The worth field is a bilinear, finite-rank, separable construction, whose three inputs are processed separately and combined only in the final
contraction. It is not a distribution over regimes. It is the scalar result of a mixture whose \emph{coefficients} are distributions over regimes. Under valid router outputs and regime gates in $[e^{-B},1],\;
e^{-B}\le W_\theta(i,j)\le1.$ This bound is a consequence of convex combination, not a normalization target. The additive attention bias is
\[
B_\theta^\ell(i,j):=\log\left(\max\{W_\theta^\ell(i,j),\varepsilon\}\right)\in[-B,0],
\]
with the default $\varepsilon=10^{-6}$ and $B=6$, one has
$\varepsilon<e^{-B}$, so the clamp is inactive for mathematically valid router
outputs. It is a numerical guard only. The bias is shared across query heads in the same layer, whereas the standard scaled similarity
$S_{ij}^{\ell,h}$ remains head-specific. Let $M_{\mathrm{causal}}(i,j)=-\infty$ if $j>i$ (0 otherwise). The complete MoSAR formula is:
\[\label{complete_MoSAR_formula}
\operatorname{Att}_{\MoSAR}(Q,K,V)
=
\softmax\left(
\frac{\widetilde Q\widetilde K^\top}{\sqrt{d_h}}
+B_\theta(\widetilde Q,\widetilde K,D)
+M_{\mathrm{causal}}
\right)V.
\]
Where \(D\footnote{\(D\) is not a learned quantity and does not
encode semantic similarity; it only provides the positional coordinate on which
the regime-dependent MoSAR geometry is evaluated.}\in\mathbb{N}_0^{n\times n}\) denotes the causal relative-distance
matrix, with \(D_{ij}=i-j\) for \(j\leq i\). Future positions \(j>i\) are handled
by the causal mask. See Appendices \ref{app:pairwise-regime-construction}, \ref{Fully expanded worth field} and \ref{Minimal pseudocode} for further details on modeling choices, the fully expanded MoSAR formula, and minimal pseudocode.

\paragraph{Training.}\label{Normalized reach cost and hard domain} MoSAR is designed to be trained as a continuous relaxation but is designed to admit
a discrete inference geometry: soft routing allows the model to learn which
interaction scales are useful, while top-1 routing converts these learned
preferences into a support that can, in principle, be determined before the
quadratic query--key similarity operation. All models are trained with the standard language-modeling loss
\(\mathcal L_{\mathrm{LM}}\), except for MoSAR-Cost, where we add an auxiliary
term to study the effect of explicit pressure toward shorter attention reach. Let $c=(0.0625,0.25,1)$ denote the normalized nominal reaches of the short, medium, and global regimes
at the training context length. We use these quantities as a proxy for the computational extent of the induced attention geometry. The role-averaged
auxiliary term is:
\begin{equation}\label{cost_term}
\mathcal L_{\mathrm{cost}}
:=
\frac12
\mathbb E_{b,\ell,i}
\left[
\pi_{b,i,Q}^{\ell}\cdot c
+
\pi_{b,i,K}^{\ell}\cdot c
\right].
\end{equation}
MoSAR-Cost is trained with
$\mathcal L_{\mathrm{LM}}
+
\lambda\mathcal L_{\mathrm{cost}},$ while the auxiliary term is disabled for all other models, including MoSAR,
for which
\(\mathcal L_{\mathrm{MoSAR}}=\mathcal L_{\mathrm{LM}}\).
The resulting expected normalized reach characterizes the geometry selected by
the routers and should not be interpreted as measured FLOPs or wall-clock
acceleration. During training, routing remains soft. Query- and key-side distributions are
combined into the continuous regime mixture defined above, and all admissible
query--key similarities are evaluated densely. This provides a differentiable
relaxation in which the routers and controlled decay can be learned jointly
with the language model. Soft MoSAR can also be used at inference when the
continuous geometry itself is desired, especially during the testing phases, but dense evaluation alone does not
reduce the number of query--key dot products.

\paragraph{Inference-time discretization.}\label{Inference-time discretization}
For compute-efficient inference, the natural projection of the learned continuous geometry is deterministic top-1 routing. For each layer, we define $m_i^{*,\ell}:=\operatorname*{arg\,max}_m\pi_{i,Q}^{\ell}(m),\;n_j^{*,\ell}:=\operatorname*{arg\,max}_n\pi_{j,K}^{\ell}(n),
$ and replace each routing distribution by its corresponding one-hot assignment. When the learned routing distributions are sufficiently concentrated, the soft worth field in Eq.~\ref{worth_field} is dominated by
\(g_{m_i^*,n_j^*}(i-j)\), so that top-1 routing is a natural discrete approximation of the continuous geometry learned during training. The discrete routing decisions also induce a computational support before pairwise query--key similarities are evaluated. For non-\(GG\) regime pairs, let $
\delta_{m_i^*,n_j^*}
=
(\delta_{m_i^*}+\delta_{n_j^*})/2.
$
We define the corresponding causal support as
$$
\Omega_\theta^\ell(x)
:=
\left\{
(i,j):
j\le i,\;
\left[
i-j\le
\delta_{m_i^{*,\ell},n_j^{*,\ell}}
\;\lor\;
\bigl(m_i^{*,\ell},n_j^{*,\ell}\bigr)=(G,G)
\right]
\right\}.
$$
The \(GG\) pair remains globally admissible, consistently with its non-decaying geometry. Crucially, \(\Omega_\theta^\ell(x)\) depends only on router outputs, regime reaches, relative distances, and validity masks; constructing the support does not require evaluating \(q_i^\top k_j\). In the dense top-1 evaluation used in this work, all causal query--key similarities are still computed and the selected regime decay is applied everywhere, including its finite floor outside the nominal reach. In a sparse implementation, instead, \(\Omega_\theta\) would be constructed first: query--key products would be evaluated only for \((i,j)\in\Omega_\theta\), the regime-dependent decay would be retained as an additive bias within the selected support, and pairs outside \(\Omega_\theta\) would be omitted from the attention computation. Thus, a distant interaction can be retained without evaluating all intermediate query--key pairs. Defining the \textit{support density} $
\rho_\Omega
:=|\Omega_\theta|/|\Omega_{\mathrm{causal}}|,
$
the similarity computation of such an implementation would scale as
$
O\!\left(
|\Omega_\theta|H_Qd_h
\right)
=
O\!\left(
\rho_\Omega n^2H_Qd_h
\right),
$
up to the cost of constructing and executing the sparse support. The exact set \(\Omega_\theta\) is a mathematical description of the desired support; practical implementations may reconstruct it using hardware-friendly sparse blocks or tiles, potentially evaluating a small superset of the selected pairs. Such an implementation reduces the quadratic constant whenever \(\rho_\Omega<1\), and becomes sub-quadratic if \(\rho_\Omega\) decreases with sequence length. The decreasing normalized routing reach observed under length extrapolation in Section~\ref{sec:results} is consistent with this possibility, although realizing and benchmarking an efficient sparse kernel is left to future work.

\section{Experimental setup}\label{Experimental setup}

We train and evaluate eight Gemma2-style~\citep{team2024gemma} 500M-parameter decoder-only language models from scratch to remain close to the experimental setting of \citet{barbero2024round}, a primary reference for this work (see also Appendix~\ref{app:backbone-choice}). We use different attention geometries designed to disentangle positional encoding, locality, smooth decay, hard truncation, and adaptive routing. \textbf{RoPE}~\citep{su2021roformer} is the dense Transformer baseline and represents the standard positional geometry used by the backbone model. \textbf{ALiBi}~\citep{press2021train} is included as a strong extrapolation-oriented baseline based on an explicit distance-dependent bias. \textbf{0.75-RoPE}~\citep{barbero2024round} provides an alternative positional geometry, allowing us to test whether length transfer can be obtained by modifying the positional encoding alone. We also include two fixed-regime MoSAR variants, \textbf{Fixed-S} and \textbf{Fixed-M}, which force all queries and keys to use the short or medium decay regime, respectively. These baselines isolate the effect of locality from adaptive routing. Finally, \textbf{RoPE-M-mask} applies a hard medium-range truncation on top of RoPE, serving as an abrupt-mask counterpart to smooth decay. The two adaptive variants are \textbf{MoSAR}, which learns routing and controlled decay without an explicit cost penalty, and \textbf{MoSAR-Cost}, which adds an expected attention-cost regularizer.

All models have 18 Transformer layers, hidden size 1024, 8 attention heads, 1 query group, FFN hidden size 8192, and a Gemma2-style query pre-attention scalar of 256. All variants use a vocabulary size of 256k and are trained with sequence length 2048, micro-batch size 1, and global batch size 8, corresponding to 16,384 training tokens per optimization step. The routers are two-layer MLPs with hidden dimension \(d_r=64\). To isolate the effect of the attention geometry, all variants share the same model initialization seed, data seed, tokenizer, packed-token stream, learning-rate schedule, precision/backend stack, and training-context evaluation protocol.

We optimize with AdamW~\citep{loshchilov2017decoupled}, using a peak learning rate of \(3 \times 10^{-4}\), a minimum learning rate of \(3 \times 10^{-5}\), and 100 warmup steps. All variants are trained with the standard language-modeling cross-entropy loss, except MoSAR-Cost, which adds the expected attention-cost term defined in~\ref{cost_term}. In our experiments, MoSAR-Cost uses \(\lambda=0.01\) with a 500-step cost warmup. The target budget is 610k optimization steps, corresponding to approximately 10B training tokens. Since the indexed English Wikipedia corpus contains approximately 5B tokens, this budget corresponds to roughly two effective passes over the corpus. Intermediate milestones at 50k, 100k, and 300k steps correspond to approximately 0.82B, 1.64B, and 4.92B training tokens, respectively.

After training, we perform a post-hoc evaluation phase to test length extrapolation and top-1 routing approximation without further parameter updates. For soft/native extrapolation, checkpoints trained at 2048 tokens are evaluated at 4096 and 8192 tokens. To keep the number of validation tokens approximately matched across extrapolation lengths, we use 64 evaluation iterations at 2048 tokens, 32 evaluation iterations at 4096 tokens, and 16 evaluation iterations at 8192 tokens, with the same global batch size of 8. For MoSAR variants, we additionally evaluate deterministic hard top-1 routing at each context length, using the same checkpoints and validation protocol.

\section{Results}\label{sec:results}

\begin{table}[t]
\centering
\small
\caption{Performance and length extrapolation in soft/native inference mode after 610k optimization steps. Expected cost is normalized with respect to full attention. 
All models are evaluated on the same validation corpus, rebuilt at each context length. Lower LM loss and perplexity are better.
}
\label{tab:main-610k}
\setlength{\tabcolsep}{4pt}
\begin{tabular}{lrrrrrrr}
\toprule
Method & LM loss & PPL@2k & PPL@4k & PPL@8k & $\Delta\%$@2k & $\Delta\%$@8k & Reach \\
\midrule
\textbf{MoSAR (ours)} & 2.5071 & 12.269 & \textbf{11.753} & \textbf{12.208} & -1.99 & -19.6 & 0.2210 \\
ALiBi & 2.5186 & 12.411 & 11.873 & 12.232 & -0.85 & -19.4 & 1.0000 \\
Fixed-S (ours) & 2.5107 & 12.313 & 11.864 & 12.453 & -1.63 & -17.9 & 0.0625 \\
Fixed-M (ours) & 2.5069 & \textbf{12.267} & 11.827 & 12.520 & -2.00 & -17.5 & 0.2500 \\
RoPE-M-mask & 2.5419 & 12.703 & 12.212 & 12.584 & +1.49 & -17.1 & 0.2500 \\
MoSAR-Cost (ours) & 2.5299 & 12.553 & 12.053 & 13.276 & +0.28 & -12.5 & 0.0897 \\
RoPE & 2.5271 & 12.517 & 12.373 & 15.176 & +0.00 & 0.0 & 1.0000 \\
0.75-RoPE & 2.5185 & 12.409 & 14.001 & 29.594 & -0.86 & +95.0 & 1.0000 \\
\bottomrule
\end{tabular}
\label{tab:main}
\end{table}

\begin{figure}[t]
    \centering
    \input{figures/training_curve_seq2048_ppl_pgfplots}
    \caption{
    Validation perplexity over pre-training checkpoints at the training context length.
    MoSAR remains competitive throughout training and reaches the final checkpoint within the strongest group of variants, while learning an adaptive controlled-decay geometry.
    }
    \label{fig:training-curve}
\end{figure}
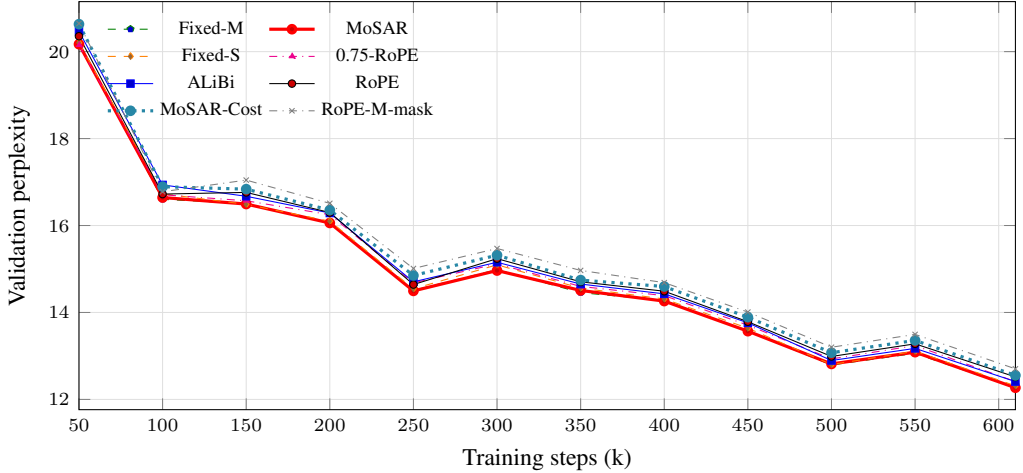

\paragraph{Learned Geometry during Pre-training.} Table~\ref{tab:main-610k} reports the training-context evaluation after 610k optimization steps. Fixed-M obtains the best perplexity, but MoSAR is essentially tied with it, differing by only 0.002 PPL while using a lower expected cost. Fixed-S also improves over dense RoPE, confirming that locality is a strong inductive bias for language modeling. However, MoSAR reaches the same quality range without committing to a single fixed scale, suggesting that the learned controlled-decay geometry can recover the benefits of locality while retaining adaptive regime selection. The comparison with RoPE-M-mask is particularly diagnostic. RoPE-M-mask has the same nominal cost as Fixed-M, but performs substantially worse, despite applying a medium-range truncation on top of RoPE. This separates smooth controlled decay from abrupt masking: the gains are not explained by reducing attention to a local window alone. MoSAR-Cost further illustrates the quality--cost trade-off: it achieves the lowest (learned) expected cost, but pays a small perplexity penalty at the training context length, suggesting that explicit cost pressure may over-compress the learned geometry. Figure~\ref{fig:training-curve} shows that the observed behavior is not a final-checkpoint artifact. Across pre-training, MoSAR remains consistently below dense RoPE and clearly separated from the hard RoPE-M-mask baseline. This supports the interpretation that controlled decay acts as an optimization and geometric regularizer throughout training, rather than merely producing a favorable endpoint after convergence. Table~\ref{tab:main-610k}
also evaluates soft/native inference under length extrapolation. MoSAR achieves the best perplexity at both 4k and 8k tokens, slightly improving over ALiBi at 8k and substantially outperforming dense RoPE. Fixed-S and Fixed-M remain strong, confirming again that locality is a powerful prior, but neither matches the adaptive MoSAR geometry at longer contexts. The severe degradation of 0.75-RoPE and dense RoPE at 8k further indicates that length transfer is not automatic: different positional geometries induce very different extrapolation behavior.

\paragraph{Discrete Approximation via Top-1 Routing.} For inference approximation, we evaluate the deterministic top-1 routing domain defined in Section~\ref{Inference-time discretization}. We refer to this deterministic discretization as hard top-1 routing. Table~\ref{tab:mosar-routing} shows that hard routing increases perplexity relative to the soft geometry, as expected, but the degradation remains bounded. For MoSAR, hard top-1 routing reduces cost increasingly with context length and remains substantially better than dense RoPE at 4k and 8k tokens. This supports the central approximation claim: the learned soft decay field can be discretized into a lower-cost support without catastrophic loss. Table~\ref{tab:mosar-routing} also provides a compact view of the learned routing behavior across context lengths. MoSAR keeps a remarkably stable query-side distribution, while key-side routing gradually shifts mass from short to medium/global regimes as longer contexts become available. This asymmetry is consistent with the role of keys as reusable memory positions: queries remain mostly local in their demand, whereas keys become increasingly available for longer-range interactions. MoSAR-Cost shows a more compressed geometry, but also increases medium/global usage at 8k, indicating that even under explicit cost pressure the model reopens longer-range regimes when extrapolating. Interestingly, under top-1 routing, the support density $\rho_\Omega$ introduced at the end of Section \ref{Inference-time discretization} decreases from 0.154 at $n{=}2048$ to 0.104 at $n{=}8192$ for MoSAR, i.e. a 33\% reduction over a 4× longer context, indicating that the effective attention cost grows sub-quadratically over the evaluated range.

\begin{table}[t]
\centering
\footnotesize
\caption{
Top-1 approximation of learned MoSAR geometries and Soft MoSAR routing statistics across context lengths (medium is omitted, since $q_M=1-q_S-q_G$, likewise for keys). 
The gap is the relative perplexity increase of discrete routing with respect to the corresponding soft geometry.
}
\label{tab:mosar-routing}
\setlength{\tabcolsep}{2.6pt}
\begin{tabular}{llrrrrrrrrr}
\toprule
Model & Length & Soft PPL & Top-1 PPL & Gap & Soft reach & Top-1 reach & $q_S$ & $q_G$ & $k_S$ & $k_G$ \\
\midrule
\textbf{MoSAR} & 2k & 12.269 & 12.703 & +3.5\% & 0.2193 & 0.1544 & 0.741 & 0.108 & 0.681 & 0.166 \\
\textbf{MoSAR} & 4k & 11.753 & 12.216 & +3.9\% & 0.1878 & 0.1276 & 0.734 & 0.114 & 0.666 & 0.180 \\
\textbf{MoSAR} & 8k & 12.208 & 12.872 & +5.4\% & 0.1690 & 0.1038 & 0.734 & 0.115 & 0.640 & 0.181 \\
\midrule
MoSAR-Cost & 2k & 12.553 & 13.123 & +4.5\% & 0.0887 & 0.0724 & 0.955 & 0.010 & 0.890 & 0.021 \\
MoSAR-Cost & 4k & 12.053 & 12.737 & +5.7\% & 0.0562 & 0.0424 & 0.934 & 0.017 & 0.878 & 0.020 \\
MoSAR-Cost & 8k & 13.276 & 13.795 & +3.9\% & 0.0858 & 0.0481 & 0.812 & 0.079 & 0.794 & 0.051 \\
\bottomrule
\end{tabular}
\label{tab:mosar-geometry}
\end{table}

\section{Limitations and Future Work}
\label{sec:limitations}

This work should be read as a controlled first study of MoSAR, rather than as a
complete evaluation of an efficient-attention system. Due to computational resource constraints, our experiments are limited in model scale, training budget, number
of seeds, and breadth of downstream evaluation. However, at this stage, the main goal is to test whether regime-based attention
geometries can preserve language-modeling quality under lower expected attention
cost, and whether the gains can be explained by locality alone. For this reason, we consider it premature to draw conclusions on complex NLP
tasks. Perplexity and training dynamics are the appropriate first signals for a
method that modifies the attention mechanism itself. A definitive chat model
would require substantially larger-scale pretraining, followed by instruction
tuning and alignment procedures such as supervised fine-tuning~\citep{hu2021lora,dettmers2023qlora} and reinforcement
learning from human feedback~\citep{ouyang2022training}, or related
preference-optimization methods~\citep{rafailov2023direct}. MoSAR should
therefore be evaluated under language-model scaling laws before claims about
downstream reasoning, instruction following, or long-context behavior can be
made~\citep{kaplan2020scaling,hoffmann2022training}. Finally, our current cost metric is an expected attention-cost proxy, not a
measured wall-clock speedup. Real acceleration will require optimized sparse
kernels, careful memory-layout design, and hardware-aware implementations, in
the spirit of IO-aware and parallel attention kernels such as FlashAttention and
FlashAttention-2~\citep{dao2022flashattention,dao2023flashattention}. Future work should scale MoSAR, evaluate long-context behavior, and develop sparse kernels.

\section{Conclusions}

The central idea of this work is that attention might not be required to decide in advance which interactions between tokens are relevant. Instead, the model can learn a computational geometry in which relevance is allowed to emerge. In this sense, MoSAR gives attention a structured space in which to become selective, adaptive, and computationally accountable. Our experiments support three main conclusions. First, controlled decay is a stronger and more stable inductive bias than abrupt local masking. Second, adaptive regime selection improves over fixed locality in the long-context setting, showing that MoSAR is not merely learning to be local. Third, the soft geometry learned by MoSAR is approximable: hard routing reduces the attention domain while preserving competitive perplexity, especially under length extrapolation.

\subsection*{AI use statement}

In this work, we used generative AI tools as general-purpose research and writing assistants. In particular, AI tools were used to support brainstorming, paper organization, language editing, LaTeX polishing, table and figure formatting, code assistance for auxiliary analysis scripts, and discussion of experimental results. We also used AI tools to help refine the presentation of the main conceptual framing, including the interpretation of MoSAR as a controlled-decay attention geometry.

We did not use generative AI tools to generate training data, fabricate or modify experimental results, run experiments, or make autonomous scientific decisions. All model implementations, training runs, evaluation scripts, numerical results, tables, and figures were inspected, executed, and verified by the authors. AI-assisted code was reviewed and tested before use, and AI-assisted text was checked against the experimental evidence and relevant literature. The authors take full responsibility for the final content of this work, including all text, claims, code, figures, tables, and artifacts produced with the aid of generative AI.

\subsection*{Reproducibility statement}

We make several efforts to support reproducibility. The MoSAR mechanism is defined in Section~\ref{mosar}, including the router architecture, regime geometry, normalized reach cost, and hard-domain construction. Section~\ref{Experimental setup} specifies the model architecture, training budget, optimization hyperparameters, dataset scale, checkpointing schedule, and evaluation protocol used for all variants. Additional theoretical, implementation, and hardware details are provided in the appendices.

The experiments are designed as controlled comparisons: all variants share the same initialization seed, data seed, tokenizer, packed-token stream, learning-rate schedule, precision/backend stack, and validation protocol, so that the attention geometry is the primary changing factor.

A repository containing the training and evaluation scripts, configuration files, dataset-preparation utilities, table- and figure-generation scripts, sanitized evaluation outputs, and detailed reproduction instructions is available at:

\url{https://github.com/michelepao1993-dev/attention-experiments.git}.

Checkpoints at 610k are available at:

\url{https://huggingface.co/michelepao1993/mosar-final-checkpoints}

To further support both understanding and reproducibility, we organize the supplementary material into two complementary blocks: Appendix \ref{app:catch22} develops the conceptual and architectural rationale behind MoSAR, in particular Appendices \ref{app:semantic-sparse-computation},\ref{app:sparse-qk},\ref{app:lemma},\ref{app:pairwise-regime-construction} discuss sparse-support computation, regime-resolution trade-offs, an approximation Lemma, and scaling with context length, while the rest of the Appendices document the notation, conventions, implementation, repository structure, and hardware details needed to reproduce and audit the work.

\bibliography{iclr2027_conference}
\bibliographystyle{iclr2027_conference}

\appendix
% \section{Appendix}
% ============================================================
% Appendix: Conceptual route to MoSAR
% ============================================================

\section{From the Catch-22 of Attention Approximation to MoSAR}
\label{app:catch22}

This appendix documents the conceptual and mathematical perspective that led to MoSAR. It is intentionally more discursive and is not meant to replace the empirical evidence in the main paper, nor does it
claim that MoSAR is theoretically optimal. Rather, it may provide useful starting points for further work on adaptive and sparse attention.

\subsection{The circularity of sparse attention}

A major structural
difficulty that any attention-approximation method must face is that, to approximate
attention, one would like to know which query--key interactions can be ignored,
but identifying such interactions is itself one of the functions of dense
attention. This view is consistent with prior theoretical evidence that sparse Transformer
patterns can preserve essential properties of full attention. For example, appropriate sparse attention patterns can retain universal
approximation properties while reducing the quadratic
memory dependence of full attention~\citep{zaheer2020big}. This circularity motivates the design choice behind MoSAR. Instead of attempting
to predict exact relevance in advance, MoSAR learns a computational regime in
which relevance is allowed to emerge. The resulting approximation is therefore
not a fixed decay rule, nor a rigid local mask, but a content-dependent geometry
over query--key interactions.

Consider a sequence of tokens \((t_n)_{n\in\mathbb{N}}\) and their corresponding
input embeddings \((x_n)_{n\in\mathbb{N}}\), with \(x_n\in\mathbb{R}^d\).
For each prefix length \(n\), let
\[
Q(n)=(q_i)_{1\leq i\leq n},\qquad
K(n)=(k_i)_{1\leq i\leq n},\qquad
V(n)=(v_i)_{1\leq i\leq n},
\]
where \(q_i,k_i,v_i\in\mathbb{R}^d\) denote the query, key, and value vectors
obtained from the model at the corresponding layer. Dense causal self-attention
is governed by the logit matrix
\[
A(n)=\frac{Q(n)K(n)^\top}{\sqrt{d}}\in\mathbb{R}^{n\times n},
\]
together with the causal mask $M$, and computes
\[
\Att(Q(n),K(n),V(n))
=
\softmax(A(n)+M)V(n).
\]
The quadratic cost of attention comes from the fact that, for each query index
\(i\), the model scores all admissible key positions \(j\leq i\).

A generic sparse approximation seeks to replace \(A(n)\) by a sparse matrix
\(S(n)\), typically with many entries set to \(-\infty\), and to compute
\[
\SparseAtt(S(n),V(n))
:=
\softmax(S(n))V(n).
\]
This gives rise to the following abstract problem.

\paragraph{Attention Approximation Problem.}
Given sequences of matrices \((Q(n))_n\), \((K(n))_n\), and \((V(n))_n\), determine
a family of sparse matrices \(S=(S(n))_n\) such that
\[
\Att(Q(n),K(n),V(n))
\approx
\SparseAtt(S(n),V(n)).
\]
The difficulty is immediate. In order to construct \(S(n)\), one must decide
which entries of \(Q(n)K(n)^\top\) are dispensable. But knowing which
query--key interactions are dispensable is precisely what dense attention is
designed to reveal. This is the catch-22 of attention approximation:
\[
\text{to avoid computing dense attention, one needs relevance;}
\]
\[
\text{to know relevance, one computes attention.}
\]
Hard local masks break this circularity by imposing a fixed rule before seeing
the data. Dense attention avoids the decision but pays quadratic cost. MoSAR
takes a different route: it learns regimes that determine where attention should
be local, where it should be medium-range, and where longer-range interactions
should remain available.

\subsection{the infinite-dimensional nature of the attention approximation problem}
\label{app:limit-not-object}

The formalism above uses sequences because attention approximation is not a
problem over one fixed matrix. For each context length \(n\), the model induces
a different matrix \(Q(n)K(n)^\top\in\mathbb{R}^{n\times n}\). As \(n\) grows,
these matrices do not belong to a single fixed finite-dimensional space. One can
embed them into an infinite two-index array, but this does not produce an
operational target that the model is trying to approximate. In natural language,
there is no final context known in advance: the next tokens depend on the user,
the task, and the continuation itself.

This matters because attention approximation cannot be defined as approximating
a pre-existing limit \(QK^\top(\infty)\). Such an object would require knowing
the completed sequence before the computation takes place. At most, one can
study conditional or pointwise notions of approximation under explicit
assumptions on the structure of the logits. But for general language modeling,
there is no universal limiting attention matrix whose sparse approximation can
be chosen a priori.

In this sense, MoSAR does not approximate a
fixed infinite object. It learns, along the sequence, an adaptive geometry: at each context length, the model decides which computational regime
is sufficient for the current query--key interaction. The data induce the representations, the representations induce
query--key similarities, and the routing mechanism learns which interactions can
be ignored, regularized, or left unchanged.

\subsection{Semantic sparsity}

The previous discussion suggests that the relevant sparsity is not merely
positional. A matrix can be sparse\footnote{For the avoidance of doubt, we refer to a sparse matrix as a matrix that can have many elements set to either 0 or $-\infty,$ e.g. $Q(n)K(n)^\top$ has many $-\infty$ $\Rightarrow$ $\softmax(Q(n)K(n)^\top)$ has many 0.} because of a fixed architectural pattern, such
as a sliding window, but the sparsity needed by language modeling is better
viewed as \textit{semantic}: most pairwise interactions are unnecessary, yet the support
of the useful interactions is not known a priori.

Let \((A(n))_n\) and \((B(n))_n\) be two sequences of matrices in
\(\mathbb{R}^{n\times d}\), with rows \(a_i(n)\) and \(b_j(n)\). Let
\[
C(n)=\big(a_i(n)^\top b_j(n)\big)_{1\leq i,j\leq n}
\]
be a sequence of similarity matrices.

\paragraph{Semantic sparsity.}
We say that \((C(n))_n\) is \textit{semantically sparse} if there exists a family of sparse
matrices \(S=(S(n))_n\) such that, in the relevant attention-induced metric,
\[
C(n)\approx S(n).
\]
In the attention setting, the object of interest is \(C(n)=Q(n)K(n)^\top\).

This definition is intentionally abstract. It does not assume that sparsity is
local, block-structured, random, or low-rank. It only states that the dense
similarity matrix may contain many interactions that are not computationally
necessary for the resulting attention output. The main question is therefore not
whether sparsity may exist, but how the model can find or induce the right sparse
geometry without first computing the full matrix.

This distinction is central to MoSAR. A hard local mask assumes that semantic
sparsity coincides with distance. MoSAR instead assumes that semantic sparsity is
regime-dependent: local structure is often sufficient, but some queries or keys
may require a larger computational reach.

\subsection{From semantic sparsity to sparse computation}
\label{app:semantic-sparse-computation}

Semantic sparsity separates two problems that are often conflated. The first is
\emph{support discovery}: determining which query--key interactions are worth
evaluating. The second is \emph{sparse computation}: once such a support is
known, computing only the corresponding entries of \(QK^\top\) without
materializing the dense product.

Suppose that, in the attention-induced sense introduced above,
$$
Q(n)K(n)^\top \approx S(n),
$$
where \(S(n)\) has support \(\Omega(n)\subseteq\{(i,j):j\leq i\}\). If
\(\Omega(n)\) were available before evaluating pairwise similarities, there
would be no need to form the complete matrix \(Q(n)K(n)^\top\). One could
instead compute
$$
\left\{
q_i^\top k_j:(i,j)\in\Omega(n)
\right\},
$$
and omit the remaining products. Since \(Q\) and \(K\) themselves remain dense
while only selected entries of their product are required, this operation is
naturally viewed as a sampled dense--dense matrix multiplication: the sparse
object is not an input matrix, but the prescribed support of the output.

This perspective clarifies the computational role of MoSAR. The routers do not
attempt to predict the exact value of \(q_i^\top k_j\). Instead, they construct
a coarse geometric domain in which evaluating that similarity is considered
potentially useful. Given deterministic routing decisions, this domain can be
constructed from token representations, regime assignments, and relative
distance without first computing the pairwise similarity matrix. In an
idealized well-trained MoSAR model, interactions outside this domain would have
sufficiently predictable negligible contribution that their exact similarities
need not be evaluated. The remaining problem is therefore algorithmic: given
a router-induced support \(\Omega_\theta(n)\), how should one compute
$$
\left(Q(n)K(n)^\top\right)\big|_{\Omega_\theta(n)}
$$
efficiently on modern hardware?

This is distinct from simply applying a mask after dense matrix
multiplication. A dense implementation first pays the \(O(n^2d)\) similarity
cost and only then removes unwanted entries. A genuinely sparse implementation
must instead use \(\Omega_\theta(n)\) as a computational stencil: first
construct the admissible domain, then evaluate \(q_i^\top k_j\) only through
the openings of that stencil, and finally apply the corresponding MoSAR
geometric bias. The current work studies whether such a stencil can be learned;
designing kernels that exploit it efficiently is a separate system problem.
In particular, an arbitrary irregular support may save arithmetic while still
mapping poorly to GPU execution, motivating block, tiled, or otherwise
hardware-aligned approximations of the learned domain.

\paragraph{Regime resolution as an approximation trade-off.}\label{app:semantic-sparse-computation}
The number of regimes \(R\) determines the resolution with which MoSAR can
describe this domain. With only a few regimes, routing is deliberately coarse:
if an interaction is considered potentially useful, the selected regime also
retains a neighborhood of nearby interactions that may ultimately be
unnecessary. This over-coverage is the price paid for deciding relevance
without evaluating every pair independently.

Increasing \(R\) provides a finer vocabulary of computational reaches and can,
in principle, tighten the approximation around useful interactions. However,
this refinement is not free. Larger regime sets increase router capacity and
routing cost, enlarge the family of regime-pair geometries, and make the
resulting support more complex to construct and execute. More fundamentally,
there is a limiting tension. If the regime system becomes sufficiently fine
that it attempts to decide relevance almost independently for every
query--key pair, the routing problem approaches the pairwise selection problem
that dense attention already solves. In this limit, the original catch-22
reappears: exact identification of all useful entries can become as difficult
as evaluating the similarities themselves.

MoSAR therefore introduces a resolution--computation trade-off:
$$
\text{coarse routing}
\quad\Longrightarrow\quad
\text{cheap support discovery but larger over-coverage},
$$
whereas
$$
\text{fine routing}
\quad\Longrightarrow\quad
\text{tighter support but greater routing and execution complexity}.
$$
Notably, bilinear routing provides non-trivial geometric resolution even with a very small regime set. For example, two marginal regimes, \(S\) and \(G\), already induce the pairwise geometries \(SS\), \(SG/GS\), and \(GG\), corresponding to short, intermediate, and global interaction scales under the symmetric construction. Thus, increasing the number of regimes refines an already combinatorial family of pairwise neighborhoods rather than merely adding one additional attention scale at a time.

% The optimal number and spacing of regimes should consequently depend on model
% scale, context length, hardware, and the statistical structure of the training
% distribution. Determining this operating point is an important direction for
% future work.

The optimal number and spacing of regimes should consequently depend on model
scale, context length, hardware, and the statistical structure of the training
distribution. Determining this operating point is an important direction for
future work. In particular, it is unclear whether the number of useful regimes
should grow proportionally with context length at all. If long-range
dependencies are organized around a small number of characteristic scales,
rather than individual token distances, a relatively small set of regimes may
remain sufficient even as the context window grows substantially. This suggests
a possible scaling law in which the number of regimes grows only slowly with
context length—for example, logarithmically under geometrically spaced
windows—or even saturates once the relevant dependency scales are covered.
Establishing whether such a relationship exists, and whether it is governed by
nominal context length or by the effective distribution of informative
dependency distances, would provide a principled basis for scaling MoSAR to
very long contexts.

\paragraph{Why longer contexts may make semantic sparsity more valuable.}
The potential benefit becomes particularly interesting as the context grows.
Suppose, schematically, that the useful support consists of a local component
of width \(w(n)\) together with \(L(n)\) genuinely useful long-range
interactions. Then
$$
|\Omega(n)|
=
O\!\left(n\,w(n)+L(n)\right),
$$
and relative to the \(O(n^2)\) causal domain its density satisfies
$$
\rho_\Omega(n)
=
O\!\left(
\frac{w(n)}{n}
+
\frac{L(n)}{n^2}
\right).
$$
Hence, if local computational reach grows sublinearly,
\(w(n)=o(n)\)\footnote{We use the standard little-$o$ notation: $f(n)=o(g(n))$ if
$f(n)/g(n)\to 0$ as $n\to+\infty$.}, and long-range dependencies remain semantically sparse,
\(L(n)=o(n^2)\), then
$$
\rho_\Omega(n)\longrightarrow 0.
$$
Under this hypothesis, longer contexts do not merely increase the absolute
benefit of sparse attention: they increase the \emph{relative} redundancy of
dense attention.

This asymptotic argument is a hypothesis rather than a result established by
the present experiments. Nevertheless, it is consistent with the behavior
observed under length extrapolation: MoSAR preserves strong language-modeling
performance while its normalized routing reach decreases as the evaluation
context grows (Table~\ref{tab:mosar-routing}). This suggests a possible scaling
picture in which much of language computation remains local while the number
of important long-range links grows substantially more slowly than the set of
all possible token pairs, and motivates the following conceptual decomposition:
\[
\text{attention computation}
\approx
\text{local semantic construction}
+
\text{sparse long-range links}.
\]
This view also clarifies why hard local attention and dense global attention
represent two unsatisfactory extremes. Hard local attention assumes that all
long-range interactions are negligible. Dense attention assumes that every
long-range interaction should remain available. MoSAR takes an intermediate
position by suppressing most distant interactions but recovering rare distant links through adaptive regimes. In this sense, MoSAR is not simply a sparse attention mechanism. It is a method
for learning when sparsity should be local, when it should be relaxed, and when
a global link may be worth its cost. In particular, MoSAR decomposes the attention approximation problem. Learned routing addresses the
question
$$
\textit{where should pairwise similarity be computed?}
$$
while sparse numerical kernels must address
$$
\textit{how should the selected similarities be computed efficiently?}
$$
The present work focuses on the first question. The second---efficiently
constructing and evaluating \(S(n)\) from a learned semantic support---is the
natural next step.

\subsection{Router-induced support and sparse query--key computation}
\label{app:sparse-qk}

The support \(\Omega_\theta\) introduced in Section~\ref{Inference-time discretization} specifies which query--key similarities are required after routing has been discretized. Its purpose is to separate a relatively inexpensive support-selection problem from the more expensive pairwise similarity computation.

For dense self-attention, evaluating all query--key products requires $
O(n^2d_h)
$ operations per head, whereas constructing or manipulating a dense \(n\times n\) Boolean support requires only \(O(n^2)\) scalar operations. A direct prototype of sparse MoSAR can therefore retain a quadratic support-construction step while avoiding the substantially more expensive \(d_h\)-dimensional dot products outside the selected domain. After top-1 routing, each query and key has a regime label,
$$
m_i^*\in\{S,M,G\},
\qquad
n_j^*\in\{S,M,G\}.
$$
The corresponding pair \((i,j)\) is retained whenever
$$
j\le i
\quad\text{and}\quad
\left[
i-j\le\delta_{m_i^*,n_j^*}
\;\lor\;
(m_i^*,n_j^*)=(G,G)
\right].
$$
Thus, one need not first materialize the complete worth field \ref{worth_field}. The router labels and relative distance are sufficient to construct a Boolean stencil for \(\Omega_\theta\). Query--key similarities are then evaluated only on this support:
$$
\{q_i^\top k_j:(i,j)\in\Omega_\theta\}.
$$
Within \(\Omega_\theta\), the corresponding MoSAR decay remains an additive geometric bias; outside \(\Omega_\theta\), the interaction is omitted from sparse attention rather than evaluated and subsequently masked. A straightforward implementation therefore has cost
$$
O(n^2)
+
O(|\Omega_\theta|d_h),
$$
instead of \(O(n^2d_h)\) for dense query--key evaluation. This already separates support discovery from similarity computation, although the support-construction step remains quadratic.

The factorized regime structure provides a path toward more efficient implementations. After top-1 routing, let
$$
Q_m^\ell:=\{i:m_i^{*,\ell}=m\},
\qquad
K_n^\ell:=\{j:n_j^{*,\ell}=n\}.
$$
Then the support can be written as the disjoint union
$$
\Omega_\theta^\ell
=
\bigsqcup_{m,n}
\Omega_{mn}^\ell,
\qquad
\Omega_{mn}^\ell
:=
(Q_m^\ell\times K_n^\ell)\cap D_{mn},
$$
where
$$
D_{mn}
:=
\{(i,j):j\le i,\; i-j\le\delta_{mn}\},
$$
with \(D_{GG}\) equal to the full causal domain. This representation separates the semantic partition induced by the routers from the geometric domain associated with each regime pair. The desired support is the union of only \(R^2\) structured pair-regime domains, with \(R=3\) in our experiments. Here, \(Q_m^\ell\) collects the query positions assigned to regime \(m\), \(K_n^\ell\) collects the key positions assigned to regime \(n\), and \(D_{mn}\) denotes the causal geometric domain allowed by the corresponding pairwise reach \(\delta_{mn}\). Thus, $
(Q_m^\ell\times K_n^\ell)\cap D_{mn}
$ contains exactly those query--key pairs whose router assignments are \((m,n)\) and whose relative distance is admissible under that pairwise geometry. An optimized implementation could therefore construct hardware-friendly block or tiled approximations
$$
\Omega_\theta
\subseteq
\widehat{\Omega}_\theta
$$
and evaluate a small superset of the exact selected pairs using block-sparse kernels. This introduces a trade-off between exact support sparsity and hardware regularity: finer supports avoid more query--key products, whereas coarser blocks may execute more efficiently on current accelerators.

This also motivates keeping the number of regimes small. The routers are intended to identify coarse regions in which interaction is plausible, not to reproduce pairwise attention scores before computing them. Too many regimes would produce increasingly fragmented supports and progressively reintroduce the original support-discovery problem. MoSAR therefore uses a small semantic partition to expose candidate ``islands'' of interaction, while the actual query--key similarity remains responsible for determining relevance inside those islands.

A useful way to interpret the potential computational gain is to compare sparse query--key evaluation with a simple dense matrix operation. While a dense \(n\times n\) matrix addition requires \(O(n^2)\) scalar operations, dense attention requires \(O(n^2d_h)\) work because each matrix entry is itself a \(d_h\)-dimensional dot product. Under a router-induced support \(\Omega_\theta\), this becomes
$$
O(|\Omega_\theta|d_h)
=
O(\rho_\Omega n^2d_h).
$$
Hence, when \(\rho_\Omega\) is sufficiently small---on the order of \(1/d_h\)---the arithmetic cost of sparse query--key evaluation approaches the scale of a simple \(O(n^2)\) matrix operation. This does not imply comparable wall-clock time, since sparse execution introduces indexing, support construction, and hardware-alignment overheads, but it highlights the central goal of MoSAR: replacing \(n^2\) expensive vector comparisons with a cheap geometric support decision followed by a much smaller set of query--key products.

\subsection{A conditional sparse-approximation lemma}\label{app:lemma}

We now state a simple sufficient condition under which sparse masking preserves
attention. The statement is deliberately conditional and simplified.\footnote{
This lemma is a streamlined version of a more general attention-approximation
argument developed during the theoretical construction of MoSAR. In the more
abstract setting, one studies \((Q(n)K(n)^\top)_n\) as a sequence of attention
matrices of increasing dimension and asks whether, as \(n\to+\infty\), full
attention can be approximated by a family of sparse attention operators. Under
suitable uniform decay assumptions on distant query--key similarities, such an
approximation can be proved. The full functional-analytic formalism is outside
the scope of this paper; nevertheless, we include this simplified argument
because it clarifies the catch-22 that MoSAR is designed to relax: sparse
approximation is justified only when the discarded tail is controlled, but such
control cannot be fixed universally for natural language.
} Its purpose is not to claim
that natural language always satisfies uniform decay, but to clarify what kind
of structure would make sparse attention mathematically legitimate.

For a fixed row \(i\), let
\[
p_i=\softmax(\alpha_i)
\]
be the dense attention distribution over admissible keys, where
\(\alpha_i=(\alpha_{ij})_j\) denotes the vector of attention logits for query \(i\).
For a window or sparse support \(\Omega_i\subseteq \{j:j\leq i\}\), define the tail mass
discarded by the support as
\[
\rho_i(\Omega_i)
=
\sum_{j\notin \Omega_i}p_{ij}.
\]
Let \(p_i^{\Omega}\) denote the attention distribution obtained by masking all
positions outside \(\Omega_i\):
\[
p_{ij}^{\Omega}
=
\begin{cases}
\dfrac{\exp(\alpha_{ij})}{\sum_{h\in \Omega_i}\exp(\alpha_{ih})}, & j\in \Omega_i,\\[1.2em]
0, & j\notin \Omega_i.
\end{cases}
\]

\paragraph{Lemma: sparse approximation under small tail mass.}
Assume that for every row \(i\),
\[
\rho_i(\Omega_i)\leq \varepsilon
\]
For some $\varepsilon>0$. Then
\[
\|p_i-p_i^{\Omega}\|_1\leq 2\varepsilon.
\]
Where \(\|\cdot\|_1\) denotes the row-wise \(L^1\) norm\footnote{Let $p\geq1$. Given $x=(x_i)_{1\leq i\leq d}\in\mathbb{R}^d, ||x||_p=\left(\sum_{i=1}^{d}|x_i|^p\right)^{1/p}.$} over the attention
distribution, while \(\|\cdot\|\) denotes any fixed norm on the value space
\(\mathbb{R}^d\).

Moreover, if the value vectors satisfy \(\|v_j\|\leq B\) for some bound $B\footnote{
The bounded-value assumption is mild in the setting considered here. At the
input level, the token embedding table is finite. At intermediate layers,
values are obtained by applying a fixed learned projection \(W_v^\ell\) to
pre-normalized hidden states. Thus, for a fixed trained model and finite
precision arithmetic, the value vectors are scale-controlled by the normalization
layer and the operator norm of \(W_v^\ell\). We use this assumption only to
translate a bound on the attention distribution into a bound on the corresponding
attention output.
}>0$, then
\[
\left\|
\sum_j p_{ij}v_j
-
\sum_j p_{ij}^{\Omega}v_j
\right\|
\leq
2B\varepsilon.
\]

\paragraph{Proof sketch.}
Let \(\Omega_i\) be the retained support and let \(\rho_i=\sum_{j\notin \Omega_i}p_{ij}\).
For \(j\in \Omega_i\), masking outside \(\Omega_i\) renormalizes the dense distribution:
\[
p_{ij}^{\Omega}=\frac{p_{ij}}{1-\rho_i}.
\]
Therefore,
\[
\sum_{j\in \Omega_i}|p_{ij}^{\Omega}-p_{ij}|
=
\sum_{j\in \Omega_i}p_{ij}
\left(
\frac{1}{1-\rho_i}-1
\right)
=
(1-\rho_i)\frac{\rho_i}{1-\rho_i}
=
\rho_i.
\]
The removed mass contributes
\[
\sum_{j\notin \Omega_i}|p_{ij}^{\Omega}-p_{ij}|
=
\sum_{j\notin \Omega_i}p_{ij}
=
\rho_i.
\]
Thus \(\|p_i-p_i^\Omega\|_1=2\rho_i\leq 2\varepsilon\). If \(\|v_j\|\leq B\), then
\[
\left\|
\sum_j(p_{ij}-p_{ij}^\Omega)v_j
\right\|
\leq
\sum_j |p_{ij}-p_{ij}^\Omega|\,\|v_j\|
\leq
B\|p_i-p_i^\Omega\|_1
\leq
2B\varepsilon.
\]
\(\square\)

% This lemma shows that sparse attention is justified whenever the discarded
% attention mass is small. A sufficient way to ensure this is a strong decay
% assumption on distant logits. For example, if there exists a monotone function
% \(g(r)\to 0\) such that
% \[
% \exp(\alpha_{ij})\leq Cg(|i-j|)
% \]
% for distant positions, then one can choose a window size \(\delta\) large enough
% so that the total mass outside \(|i-j|\leq\delta\) is small. In that case, the
% sparse approximant can be constructed from an a priori decay estimate. However, this sufficient condition exposes the limitation of fixed sparse
% attention: to know the right \(\delta\), one needs a reliable a priori estimate
% of how similarities decay. Without such an estimate, the sparse matrix \(S(n)\)
% exists only abstractly; it is not constructively available.

This lemma shows that sparse attention is justified whenever the discarded
attention mass is small. A sufficient way to ensure this is a strong decay
assumption on distant logits. For example, suppose that there exists a
non-negative monotone function \(g\) with summable tail,
$$
\sum_{r>\delta} g(r)\longrightarrow 0
\qquad\text{as }\delta\to+\infty,
$$
such that, for distant positions,
$$
\exp(\alpha_{ij})\leq Cg(|i-j|).
$$
Assume moreover that the softmax normalizing constants
$$
Z_i:=\sum_{h\leq i}\exp(\alpha_{ih})
$$
are uniformly bounded below by some \(z_0>0\). Then the attention mass outside a
window of radius \(\delta\) satisfies
$$
\rho_i(\delta)
\leq
\frac{C}{z_0}
\sum_{r>\delta}g(r),
$$
and can therefore be made arbitrarily small by choosing \(\delta\) sufficiently
large. In that case, the sparse approximant can be constructed from an
a priori decay estimate.  Without such an estimate, the existence of a low-error sparse approximation
does not by itself provide a constructive rule for identifying its support. The assumption above is naturally satisfied by the class of decaying kernels used to define the MoSAR regimes, provided the finite numerical floor is omitted. For example, a kernel of the form
\begin{equation}\label{g_r}
g(r)=\exp\!\left[-B\left(\frac{r-T}{L}\right)^p\right],
\qquad r>T,\qquad p>1,
\end{equation}
has a rapidly vanishing tail and therefore admits arbitrarily small truncation error by choosing a sufficiently large support radius. Thus, distance decay provides a direct geometric mechanism through which dense attention can become approximable.

The remaining difficulty is that natural language does not obey a single universal decay scale. Most interactions may be local, while rare long-range dependencies can remain essential; importantly, the existence of such a dependency does not imply that all intermediate positions should also be attended to. The relevant structure is therefore closer to a sparse set of long-range links than to a dense bridge over the entire context. This motivates a weaker and more realistic hypothesis: language attention may be predominantly local while remaining semantically sparse at longer distances. The approximation problem is then not to choose between local and global attention once and for all, but to identify which geometric regime is appropriate for each interaction. MoSAR addresses this by combining controlled decay with input-dependent routing, preserving the approximability induced by locality while allowing selected interactions to access broader context.

\paragraph{Remark on the choice of norm.}
The bound above is stated in \(L^1\) over each attention row, equivalently in
total variation up to a constant factor, because the rows of the attention
matrix are probability distributions. This choice is natural for controlling the
mass removed by masking. However, the choice of norm is not merely cosmetic in
the present setting. For each fixed context length \(n\), all norms on
\(\mathbb{R}^n\) are equivalent; nevertheless, the equivalence constants may
depend on \(n\). Since attention approximation is a growing-dimensional problem,
with \(Q(n)K(n)^\top\in\mathbb{R}^{n\times n}\), norm-dependent constants may
change with the context length. Thus, extending the approximation guarantee to a
generic norm requires specifying whether the associated constants are uniform in
\(n\). As anticipated in Section \ref{app:limit-not-object}, this is one reason why attention approximation is more delicate than a standard
finite-dimensional perturbation argument. For each fixed context length \(n\),
\(Q(n)K(n)^\top\) belongs to \(\mathbb{R}^{n\times n}\), but the sequence
\((Q(n)K(n)^\top)_n\) does not belong to any fixed finite-dimensional space.
Studying approximation uniformly in \(n\) therefore requires embedding these
growing matrices into an infinite-dimensional space of two-index arrays, or at
least treating the problem asymptotically as a sequence of finite-dimensional
problems with dimension tending to infinity. This distinction is standard in \textit{functional analysis}\footnote{
Functional analysis provides the mathematical language for studying spaces of
functions, operators, norms, and convergence. In our setting, it clarifies why
attention approximation over matrices of increasing size cannot be treated as a
single finite-dimensional perturbation problem.
}: norm equivalence is a finite-dimensional property, whereas uniform control
over sequences of growing-dimensional spaces requires additional structure
\citep{brezis2011functional,conway1990course}.

\subsection{From controlled decay to adaptive regimes}\label{app:pairwise-regime-construction}

\paragraph{Controlled decay.} Before MoSAR, one natural response to the approximation problem was to design a
positional or similarity mechanism with explicit, controllable decay. Let
\(g:[0,+\infty)\to[0,1]\) be a monotone decay function with \(g(0)=1\) and
\(g(r)\to0\) as \(r\to\infty\). A decay-modulated attention logit can be written
abstractly as
\[
\alpha_{ij}^{g}
=
\alpha_{ij}+\log g(|i-j|),
\]
or, in a multiplicative view over similarities,
\[
\exp(\alpha_{ij}^{g})
=
\exp(\alpha_{ij})g(|i-j|).
\]
If \(g\) is known and admits a priori bounds, then for a prescribed tolerance
\(\varepsilon\) one may solve
\[
Cg(\delta)<\varepsilon
\]
and obtain a cutoff scale \(\delta\). This makes the sparse approximant
constructive. For instance, in \ref{g_r}, a decay threshold gives
\[
\delta_\varepsilon
=
T+L\sqrt{-\log \varepsilon}
\]
up to constants. The parameters \(T\) and \(L\) have intuitive roles: \(T\) is a
plateau over which the original geometry is preserved, and \(L\) controls the
transition scale over which distant interactions are gradually suppressed. This line of reasoning is useful because it shows that sparse approximation is
not merely an implementation trick. It depends on a trade-off between three quantities: locality, regularity, approximation error.

A hard mask maximizes locality but introduces a discontinuity. Dense attention
maximizes availability but pays quadratic cost. A smooth decay attempts to
preserve the geometry of short-range attention while gradually suppressing
distant interactions. This also explains why hard locality can be brittle. Abruptly setting logits to
\(-\infty\) is not just a computational choice; it changes the geometry seen by
the softmax. In contrast, a regime-based mechanism can preserve a smoother
computational structure inside each selected regime. This is consistent with
our empirical observation that fixed MoSAR regimes outperform hard RoPE masks at
matched nominal cost.
\begin{figure}[t]
    \centering
    \includegraphics[width=0.65\linewidth]{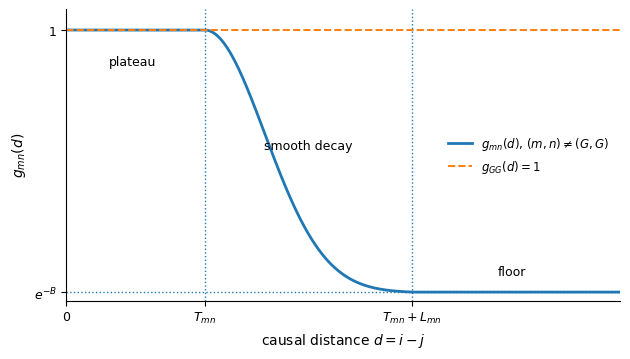}
    \caption{
    Illustration of the pairwise MoSAR decay kernel $g_{mn}(d)$ as a function
    of causal distance $d=i-j$. Interactions remain unattenuated up to the
    pair-dependent threshold $T_{mn}$, decay smoothly over a transition region
    of length $L_{mn}$, and reach the finite floor $e^{-B}$ thereafter.
    The global--global regime is exempt from decay, with $g_{GG}(d)=1$ for all
    $d\geq0$.
    }
    \label{fig:mosar-decay}
\end{figure}
The limitation of a single decay function is that it still imposes one global
notion of distance. A fixed \(g\) assumes that the same decay law is appropriate
for all tokens, layers, contexts, and tasks. MoSAR relaxes this assumption. It
does not choose one decay scale for the whole model. It lets the model choose
among a family of averaged attention regimes $(g_{mn})_{(m,n)\in\mathcal{R}\times\mathcal{R}}$ (Figure \ref{fig:mosar-decay}):
\[
g_{mn}(d):=
\begin{cases}
1, & d\le T_{mn},\\[3pt]
\exp\left[-B\left(\dfrac{d-T_{mn}}{L_{mn}}\right)^p\right], & T_{mn}<d<T_{mn}+L_{mn},\\[9pt]
e^{-B}, & d\ge T_{mn}+L_{mn}.
\end{cases}
\]
We set \(p=2\) to obtain a smooth onset of the decay: the quadratic exponent gives zero slope at \(d=T_{mn}\), continuously matching the preceding unit plateau and avoiding an abrupt change in attenuation. We set \(B=6\), corresponding to a minimum multiplicative weight \(e^{-6}\approx2.5\times10^{-3}\) and hence a maximum additive penalty of \(-6\). This provides strong finite attenuation while deliberately remaining softer than a near-hard mask; larger values such as \(B=8\) would further suppress the tail without changing the qualitative geometry.

\paragraph{Decay strength and sparse approximation.} The choice of \(B\) also determines how closely the continuous training geometry can approximate a genuinely sparse inference geometry. Consider a top-1 routed query \(i\), and let \(\Omega_i\) denote the support induced by the selected regimes. In the dense formulation, interactions outside \(\Omega_i\) are not removed: once the corresponding decay reaches its floor, their logits still take the form
$$
\alpha_{ij}^g=\alpha_{ij}-B,
\qquad j\notin\Omega_i,
$$
where, as above, \(\alpha_{ij}\) denotes the content-dependent query--key score. A sparse implementation would instead omit these similarities entirely and renormalize attention over \(\Omega_i\). Hence, the relevant quantity is not the pointwise value \(e^{-B}\) alone, but the total attention probability assigned outside the retained support,
$$
\rho_i
:=
\sum_{j\notin\Omega_i} p_{ij}.
$$
By the sparse-approximation lemma, if \(\rho_i\le\varepsilon\), then restricting and renormalizing attention on \(\Omega_i\) changes the attention distribution by at most \(2\varepsilon\) in \(L^1\). Thus, the desired transition from dense MoSAR to sparse execution is accurate whenever the learned geometry makes the discarded mass small.

Increasing \(B\) suppresses the tail more strongly and can therefore help reduce \(\rho_i\), but the relationship is not monotone in model quality. A larger \(B\) also makes routing mistakes increasingly costly during training and progressively turns the smooth decay into an approximation of a hard mask, weakening the regularizing role of the semantic transition zone. Conversely, a moderate \(B\) preserves a smoother optimization landscape and allows strong content similarity to overcome geometric attenuation, but may leave non-negligible cumulative probability outside the support. This distinction becomes especially relevant at long context lengths: although \(e^{-6}\approx2.5\times10^{-3}\) is small pointwise, many such interactions can collectively carry non-negligible mass after softmax normalization.

Our fixed-regime baselines provide useful, although indirect, evidence for this trade-off. At the training length \(n=2048\), Fixed-S and Fixed-M achieve perplexities of \(12.313\) and \(12.267\), respectively, compared with \(12.269\) for MoSAR, showing that strongly local geometries can retain most of the language-modeling quality in this regime. Under extrapolation, however, MoSAR becomes increasingly favorable: at \(8192\) tokens it reaches \(12.208\), compared with \(12.453\) for Fixed-S and \(12.520\) for Fixed-M (Table \ref{tab:main-610k}). These results are consistent with a picture in which most interactions can be handled locally, while a smaller set of longer-range dependencies becomes increasingly important as context grows. They should not be interpreted as direct measurements of discarded attention mass, since each fixed baseline is trained under its own geometry and can adapt its internal representations accordingly.

We therefore regard \(B\) as controlling a trade-off between \emph{smooth geometric regularization} and \emph{sparse approximability}, rather than as a purely numerical decay constant. The present experiments use \(B=6\) throughout and do not establish an optimal value. A more complete study should measure the discarded mass \(\rho_i\) directly under top-1 routing and evaluate how it changes with \(B\), context length, model scale, and training progress. Such measurements would connect the theoretical approximation guarantee more directly to inference-time sparsification and clarify whether stronger decay, scheduled increases of \(B\), or explicit penalties on discarded mass provide the best path from continuous training geometry to accurate sparse execution.

\begin{figure}[t]
    \centering
    \begin{subfigure}{0.235\linewidth}
        \centering
        \includegraphics[width=\linewidth]{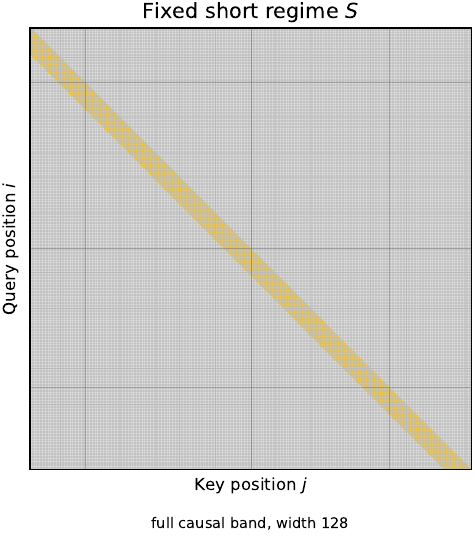}
        \caption{Fixed \(S\)}
    \end{subfigure}
    \hfill
    \begin{subfigure}{0.235\linewidth}
        \centering
        \includegraphics[width=\linewidth]{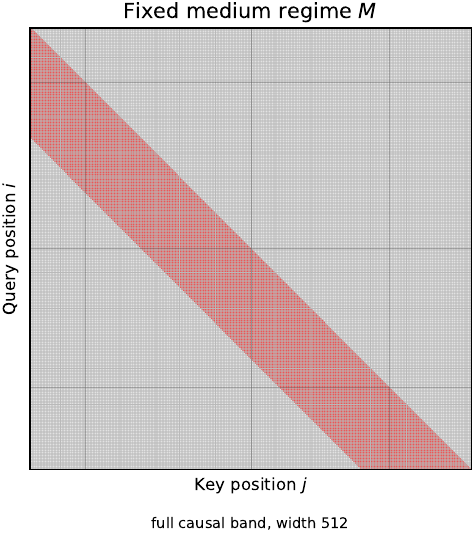}
        \caption{Fixed \(M\)}
    \end{subfigure}
    \hfill
    \begin{subfigure}{0.235\linewidth}
        \centering
        \includegraphics[width=\linewidth]{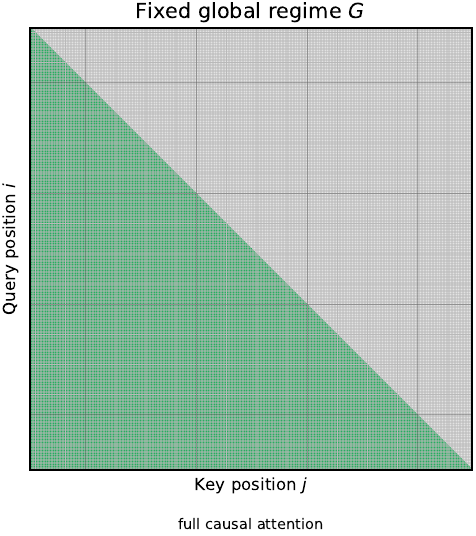}
        \caption{Fixed \(G\)}
    \end{subfigure}
    \hfill
    \begin{subfigure}{0.235\linewidth}
        \centering
        \includegraphics[width=\linewidth]{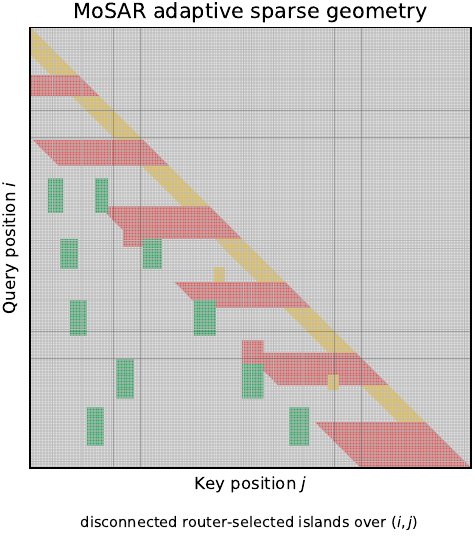}
        \caption{MoSAR}
    \end{subfigure}
    \caption{
    Schematic attention regimes at sequence length \(2048\), rendered as explicit
    cell grids so that each square corresponds to a single \(q_i k_j\) interaction.
    Fixed \(S\), \(M\), and \(G\) produce contiguous causal bands. In contrast,
    MoSAR activates only router-selected \((i,j)\) regions: the resulting pattern is sparse
    and disconnected, with isolated attention islands rather than continuous bridges across
    the entire past context.
    }
    \label{fig:mosar-regime-geometries-grid}
\end{figure}

\paragraph{Adaptive regimes.} Ultimately, MoSAR can be viewed as a learned relaxation of the fixed-decay picture. Instead
of defining one global sparse approximant \(S(n)\), MoSAR parameterizes a family
of possible attention geometries through a small set of regimes (Figure \ref{fig:mosar-regime-geometries-grid}). In the setting
used in this work, the regimes are
\[
\mathcal{R}=\{S,M,G\},
\]
corresponding to short, medium, and global computational reach. For sequence
length \(2048\), the associated normalized costs are
\[
c_S=\frac{128}{2048}=0.0625,\qquad
c_M=\frac{512}{2048}=0.25,\qquad
c_G=1.
\]
The important point is that MoSAR does not predict the exact relevant interaction in
advance. Instead, it predicts the computational regime in which the interaction
should be evaluated. This distinction breaks the catch-22 in a weaker but more
constructive way because predicting exact relevance is too hard before attention,
whereas predicting computational regime is learnable from contextual states. The router does not need to know which single token will matter. It only needs
to decide whether a query--key interaction should be treated as short-range,
medium-range, or potentially global.

This gives a direct interpretation of the cost objective used in MoSAR-Cost. The training objective is
\[
\mathcal{L}
=
\mathcal{L}_{LM}
+
\lambda\mathcal{L}_{cost}.
\]
The role of \(\mathcal{L}_{cost}\) is not to force the model into one local mask.
It biases the model toward parsimonious geometries while preserving the ability
to spend more attention budget when the contextual state demands it.

As described in Section \ref{mosar}, MoSAR must assign a geometric regime to each query--key pair \((i,j)\) before evaluating the corresponding similarity \(q_i^\top k_j\). Since the relevance of a pair is not available a priori, the model instead constructs its interaction geometry from two independently computed routing decisions: one associated with the query state \(q_i\), and one with the key state \(k_j\). Each router outputs a distribution over the same set of marginal regimes,$
\pi_{i,Q},\;\pi_{j,K}\in\Delta^{R-1},
$ with \(R=3\) in our experiments, corresponding to short, medium, and global interaction scales. We model the joint pairwise regime distribution through the factorized product
$
P_{ij}(m,n)
=
\pi_{i,Q}(m)\pi_{j,K}(n),
$
which yields \(R^2\) possible query--key regime pairs. For \(R=3\), this gives the nine combinations
$$
SS,\;SM,\;SG,\;MS,\;MM,\;MG,\;GS,\;GM,\;GG.
$$
This construction follows the natural assumption that query- and key-side routing decisions are produced independently, conditioned on their respective contextual representations, and are combined only when forming the geometry of the pair. It also avoids introducing an additional pairwise router, which would require direct access to both \(q_i\) and \(k_j\) and would partially reintroduce the pairwise computation that MoSAR is intended to anticipate.

The remaining question is how two marginal regimes should determine the geometry of their pair. Let \(\delta_m\) and \(\delta_n\) denote their nominal reaches. We use the symmetric additive composition
$$
\delta_{mn}
=
\frac{\delta_m+\delta_n}{2},
$$
and analogously combine the corresponding decay parameters. This choice is deliberately simple. It is symmetric with respect to query and key roles, preserves the original regime when both sides agree, and places mixed pairs at an intermediate scale:
$$
\delta_{mm}=\delta_m,
\qquad
\delta_{mn}\in
[\min(\delta_m,\delta_n),\max(\delta_m,\delta_n)].
$$
Thus, neither side unilaterally determines the pairwise reach. Instead, the resulting geometry can be interpreted as a compromise between the interaction scales independently proposed by the query and key representations.

Other composition rules are possible. For example, \(\min(\delta_m,\delta_n)\) would require both sides to support a long interaction, whereas \(\max(\delta_m,\delta_n)\) would allow either side to extend the pairwise reach. Such choices encode different geometric priors. We adopt the arithmetic mean because it introduces the weakest additional asymmetry or logical constraint: it simply interpolates between the two independently selected scales and leaves the language-modeling objective free to adapt the routing behavior to this geometry.

This factorized construction is central to the role of MoSAR. For a potentially useful pair \((i,j)\), the query and key routers need not predict the exact similarity \(q_i^\top k_j\); they only need to select marginal regimes whose composition assigns sufficient geometric support to that pair. The subsequent attention similarity then determines whether the interaction is actually useful. The model is therefore not instructed how to solve next-token prediction through a fixed routing semantics. Rather, it is given a small family of geometric primitives and learns, through the language-modeling objective, how to use them.

\subsection{Fully expanded formula}\label{Fully expanded worth field}

The complete MoSAR logit defined in \ref{complete_MoSAR_formula} is:
\begin{align}
A_{ij}^{\ell,h}
&=
S_{ij}^{\ell,h}
+B_\theta^\ell(i,j)
+M_{\mathrm{causal}}(i,j)
\\
&=
\frac{
(\widetilde q_i^{\ell,h})^\top
\widetilde k_j^{\ell,\gamma(h)}
}{\sqrt{d_h}}
+
\log\left(
\pi_{i,Q}^{\ell\top}
\mathbf G(i-j)
\pi_{j,K}^\ell
\right)
+
M_{\mathrm{causal}}(i,j)
.
\end{align}
The attention probabilities and output are $a_{ij}^{\ell,h}=\softmax_j(A_{ij}^{\ell,h}),$ and
$o_i^{\ell,h}=
\sum_j a_{ij}^{\ell,h}v_j^{\ell,\gamma(h)}.$ 

For causal positions, $a_{ij}^{\ell,h}
=
\frac{
\exp(S_{ij}^{\ell,h})W_\theta^\ell(i,j)
}{
\sum_{t\le i}
\exp(S_{it}^{\ell,h})W_\theta^\ell(i,t)
}.$ Thus the positive worth gate modulates the unnormalized attention weights. It would be incorrect to compute $\softmax(S)\odot W$ without renormalization. 

Suppressing the layer superscript for readability, the complete worth field is:
\begin{align}
W_\theta(i,j)
&=
\softmax\!\left(
\frac{
W_{2,Q}\GELU\!\left(W_{1,Q}\vecop(\widetilde Q_i)+b_{1,Q}\right)+b_{2,Q}
}{\tau}
\right)^\top
\mathbf G(i-j)
\\
&\quad\cdot
\softmax\!\left(
\frac{
W_{2,K}\GELU\!\left(W_{1,K}\vecop(\widetilde K_j)+b_{1,K}\right)+b_{2,K}
}{\tau}
\right).
\end{align}
The corresponding attention logit is:
\begin{align}
A_{ij}^{h}
&=
\frac{(\widetilde q_i^{h})^\top\widetilde k_j^{\gamma(h)}}{\sqrt{d_h}}
+M_{\mathrm{causal}}(i,j)
\\
&\quad+
\log\!\Bigg[
\softmax\!\left(
\frac{
W_{2,Q}\GELU\!\left(W_{1,Q}\vecop(\widetilde Q_i)+b_{1,Q}\right)+b_{2,Q}
}{\tau}
\right)^\top
\mathbf G(i-j)
\\
&\hspace{4.3cm}\cdot
\softmax\!\left(
\frac{
W_{2,K}\GELU\!\left(W_{1,K}\vecop(\widetilde K_j)+b_{1,K}\right)+b_{2,K}
}{\tau}
\right)
\Bigg].
\end{align}
This expansion makes explicit that the router biases occur inside their MLP
logits and before the regime softmaxes, while the worth gate enters attention
through an outer logarithm.

\subsection{Minimal pseudocode}
\label{Minimal pseudocode}
\begin{lstlisting}[language=Python]
# Adapter-owned operations
q, k, v = project_qkv(norm(hidden_states))
q_post, k_post = apply_positional_transform(q, k, positions)

# Model-agnostic MoSAR core
q_state = query_router(q_post)  # final softmax -> simplex
k_state = key_router(k_post)    # final softmax -> simplex

G = regime_kernel(relative_distances)
worth = einsum("bim,ijmn,bjn->bij",
               q_state.probabilities,
               G,
               k_state.probabilities)
log_worth_bias = log(clamp_min(worth, epsilon))

# Backbone/reference attention
scores = grouped_scaled_qk(q_post, k_post)
scores = scores + log_worth_bias[:, None, :, :]
scores = scores + causal_and_padding_mask
attention = softmax(scores, dim=-1)
output = grouped_attention_value_product(attention, v)
\end{lstlisting}

\subsection{MoSAR architecture diagram}
\label{app:mosar-architecture}

\begin{figure}[t]
    \centering
    \includegraphics[width=0.55\textwidth]{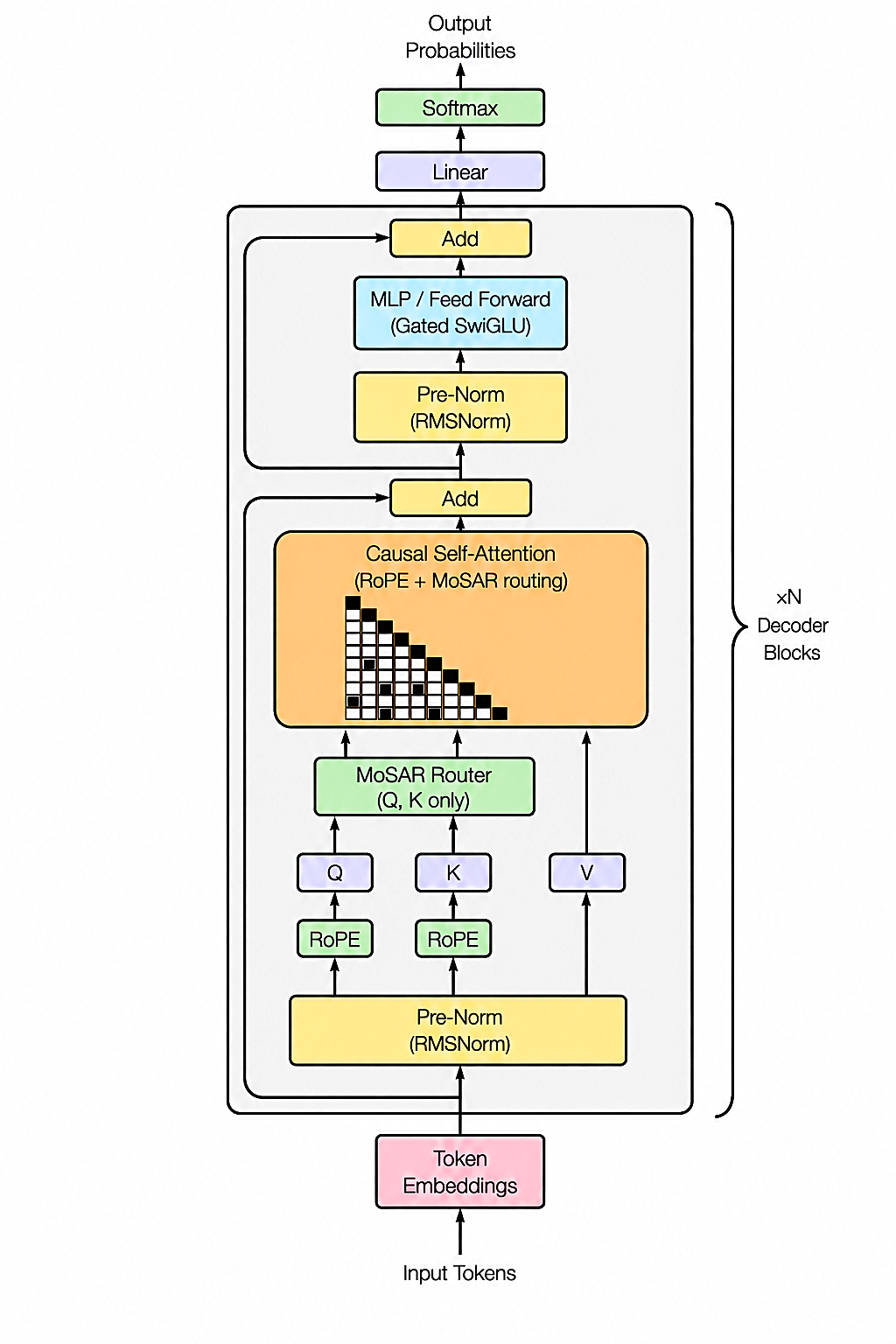}
    \caption{MoSAR-augmented decoder-only Transformer architecture. The model follows a pre-norm decoder-only backbone, while RoPE is applied internally at each layer to queries and keys rather than added once at the input level. The MoSAR router operates only on the query--key pathway and selects the attention regime used by causal self-attention, while values $V$ bypass the router. The overall layout is intentionally designed as a visual adaptation of the iconic Transformer schematic introduced by~\citet{vaswani2017attention}, so as to make the architectural differences immediately comparable. The present figure is nevertheless a new, task-specific reformulation for Gemma-style decoder-only models with post-RoPE routing.}
    \label{fig:mosar-architecture}
\end{figure}

Figure~\ref{fig:mosar-architecture} provides a full diagram of the decoder-only architecture used in this work. The figure is intended as an implementation-oriented complement to the mathematical description in the main paper. In particular, it makes explicit three design choices that are central to MoSAR: (i) the model is decoder-only and uses pre-normalization, following modern Gemma-style Transformer practice; (ii) positional information is not injected once at the input level, but instead enters each layer through RoPE applied to queries and keys; and (iii) the MoSAR router acts only on the query--key pathway, i.e., it receives information from $Q$ and $K$ but not from $V$, and modulates the attention geometry by selecting the computational regime.

More precisely, each decoder block first applies a pre-norm transformation, then computes $Q$, $K$, and $V$. RoPE is applied to $Q$ and $K$, after which the router determines the attention regime used by causal self-attention. The resulting attention output is added through a residual connection, followed by a second pre-norm, an MLP block, and a second residual connection. This diagram is useful mainly as a visual summary of the architectural role of the router and of the fact that MoSAR modifies the geometry of attention without altering the overall decoder-only backbone.

\subsection{Choice of backbone and experimental scale}
\label{app:backbone-choice}

We adopt a Gemma2-style backbone to remain close to the experimental setting of \citet{barbero2024round}, a primary reference for our study; following the Gemma2 attention formulation, logits use the backbone-specific query pre-attention scaling and \(\tanh\) soft-capping before the MoSAR bias is applied. All variants use the same Gemma2-style backbone and differ only in the positional/attention mechanism under study; accordingly, table labels such as RoPE, ALiBi, and MoSAR denote the corresponding attention variant rather than different backbone architectures.

In the article of \citet{barbero2024round}, they challenge the interpretation of RoPE as inducing a reliable distance-decay law and evaluate RoPE, NoPE, and \(p\)-RoPE using Gemma 2B models trained from scratch on Wiki and FlanV2. Our original experimental design aimed to follow this setting more closely, including 2B-parameter models and both training corpora. However, reproducing multiple matched pre-training trajectories at that scale was beyond our available computational budget.

We therefore preserve the architectural connection to Gemma and the direct \(p\)-RoPE comparison while reducing the model scale and training setting to make a controlled multi-variant study feasible. This choice prioritizes matched initialization, data order, optimization, and training budget across attention geometries over reproducing the original model scale. The resulting experiments should consequently be interpreted as a controlled mechanistic study of attention geometry rather than a scale-matched reproduction of the Gemma-2B experiments.

The experimental setting is intentionally controlled rather than large-scale. Our 500M-parameter models are trained for approximately 10B tokens, corresponding to roughly 20 training tokens per parameter and therefore to a compute regime consistent with standard scaling-law recommendations~\citep{kaplan2020scaling}. This allows us to compare multiple attention geometries under matched initialization, data order, optimization, and training budget rather than under heterogeneous pre-training conditions.

The present study focuses on whether MoSAR can learn an adaptive and approximable attention geometry, and evaluates this through language-modeling quality, routing behavior, expected normalized reach, and length extrapolation. Dedicated retrieval and long-context reasoning benchmarks would provide a complementary test of whether the learned global regimes preserve task-relevant distant information, and constitute an important direction for future work.

\subsection{Summary}

The conceptual argument above leads to a concrete empirical question:
\[
\text{Are MoSAR gains explained by locality alone?}
\]
If the answer were yes, then a hard local RoPE mask with the same nominal cost
should perform similarly to a fixed MoSAR regime. Our ablations test precisely
this possibility. The observed pattern is that hard local masks are substantially worse than fixed
MoSAR regimes at matched nominal cost (Table \ref{tab:main-610k}). This supports the claim that the regime
geometry matters beyond locality itself. In the language of this appendix, the
advantage does not come from simply deleting distant entries of \(QK^\top\). It
comes from inducing a structured sparse geometry in which the model can still
organize attention inside the selected regime.

The learned MoSAR variants address a second question:
\[
\text{Can the model learn when to spend attention budget?}
\]
MoSAR tests whether adaptive regimes are useful under language-modeling loss
alone. MoSAR-Cost tests whether explicit compute pressure can shape the learned
geometry without collapsing performance. The fact that explicit compute pressure induces a clear quality–cost trade-off suggests that compute pressure can act as a geometric regularizer,
rather than merely as a compression penalty.

A useful way to summarize the motivation behind this work is the following principle:
\begin{center}
\textit{The cost of addressing should scale with the number of relevant
destinations, \\not with the size of the entire context.}
\end{center}
Dense attention violates this principle by scoring all admissible destinations.
A hard local mask enforces it too aggressively by assuming that the only relevant
destinations lie within a fixed neighborhood. MoSAR approximates the principle
by making addressing regime-dependent.

This distinction is especially important for long contexts. If a document is
long because it contains many unrelated or weakly related regions, dense
attention pays for all pairwise interactions even when only a few are useful. If
a document contains rare but important long-range dependencies, a purely local
mask cannot recover them. MoSAR is designed for the intermediate case in which
the context is mostly locally meaningful but globally linked by sparse
relations. This is also why expected cost, rather than wall-clock speedup, is the correct
object of study in this work. We do not claim that the present implementation
achieves optimized sparse-kernel acceleration. Instead, we test whether the
learned attention geometry is compatible with lower computational reach while
preserving language modeling performance. In other words, the present
experiments evaluate whether the model learns an approximable geometry.

This appendix therefore provides a conceptual route:
\[
QK^\top
\quad\Longrightarrow\quad
\text{catch-22}
\quad\Longrightarrow\quad
\text{semantic sparsity}
\quad\Longrightarrow
\]
\[
\text{conditional sparse approximation}
\quad\Longrightarrow\quad
\text{adaptive regimes}\quad\Longrightarrow\quad
\text{MoSAR}.
\]
The empirical study in the main paper is intentionally controlled around this
route. It does not attempt to benchmark every efficient-attention alternative.
Rather, it tests the mechanistic hypothesis suggested by the theory: locality
alone is insufficient, but regime-based sparse geometry can preserve language
modeling quality under lower expected attention cost. Our results should therefore be read as a controlled mechanistic study rather
than as a broad benchmark claim. This is in line with recent work emphasizing
that high-level model behavior often reflects structural properties of the
training and evaluation process, not merely raw model capability~\citep{kalai2025language}.

The central lesson is that attention approximation should not be framed as the
problem of imposing a sparse mask, but as the problem of learning an
approximable geometry. Fixed local masks make the approximation problem easy but
brittle. Dense attention avoids the approximation but pays quadratic cost.
Fixed decay provides a mathematical bridge, but assumes a universal distance law
that natural language does not satisfy uniformly. MoSAR arises from relaxing that assumption. It replaces a single global decay
scale with content-dependent regimes. This allows attention to remain local when
locality is sufficient, expand to medium range when needed, and recover sparse
long-range links without turning them into dense bridges across the entire
context.

In short, MoSAR does not try to know relevance before attention. It learns where
attention should be allowed to look, at what scale, and at what expected cost.

\section{Notation and conventions}
\label{notation}

\begin{center}
\begin{tabularx}{\textwidth}{>{\bfseries}l X}
\toprule
Symbol & Meaning \\
\midrule
$\ell$ & transformer layer index \\
$b$ & batch index \\
$i$ & query-token index \\
$j$ & key-token index \\
$h$ & query-head index, $h\in\{1,\ldots,H_Q\}$ \\
$r$ & key/value-head index, $r\in\{1,\ldots,H_{KV}\}$ \\
$m,n$ & query and key regime indices, in $\{1,\ldots,R\}$ \\
$d_h$ & attention head dimension \\
$d_r$ & router hidden dimension \\
$N$ & configured maximum sequence length \\
$R$ & number of regimes; the default is $R=3$ \\
$\DeltaSimplex$ & probability simplex over the $R$ attention regimes\\
$\mathbb{N}$ & Set of positive integer numbers\\
$\mathbb{N}_0$ & $\mathbb{N}\cup\{0\}$\\
\bottomrule
\end{tabularx}
\end{center}

\section{Numerical semantics}

\subsection{Precision}

The repository accumulates the small regime contraction and dense reference
QK similarities in float32. Router modules otherwise follow their input/module
dtype. This is intended to improve BF16 safety while preserving gradients.

\subsection{Future positions}

\code{RegimeKernel} clamps distances below zero to zero so that the kernel is
finite for all matrix entries. This does not grant access to future positions:
the causal mask is applied separately and sets future logits to $-\infty$.

\subsection{Exact global identity}

The global-global kernel is assigned exactly one after exponentiation:

\begin{lstlisting}[language=Python]
kernel[..., global_index, global_index] = 1.0
\end{lstlisting}

Together with \code{log(clamp\_min(W, epsilon))}, this yields exact zero
log-bias for a pure global-global mixture. This supports a strict no-op
identity test against vanilla attention.

\section{Repository traceability matrix}

\begingroup\small
\begin{longtable}{p{4cm}p{4.2cm}p{4.5cm}}
\toprule
Mathematical object & Implementation & Compatibility condition \\
\midrule
\endfirsthead
\toprule
Mathematical object & Implementation & Compatibility condition \\
\midrule
\endhead

Configuration $(N,\delta,\alpha,d_r,\tau,B,p,\varepsilon)$
& \codepath{mosar/config.py}: \codepath{MoSARConfig}
& Values and validation must match the experimental configuration.\newline \\

$x_{i,Q},x_{j,K}$
& \codepath{mosar/routers.py}: \codepath{PostPositionRouter.forward}
& Inputs must be post positional; key heads must be unexpanded GQA heads.\newline \\

$u_Q,z_Q,\pi_Q$ and $u_K,z_K,\pi_K$
& \codepath{mosar/routers.py}
& Separate Q and K parameters; GELU MLP; final temperature scaled softmax.\newline \\

Router initialization
&
\codepath{mosar/routers.py}:
\codepath{PostPositionRouter.reset_parameters}\newline
& $W_1$ std $0.02$, $W_2$ std $10^{-3}$, all biases zero.\newline \\

$\delta_{mn},T_{mn},L_{mn}$
& \codepath{mosar/regimes.py}: \codepath{pair_geometry}
& Arithmetic means and transition definitions must match the equations.\newline \\

$\mathbf G(d)$
&
\codepath{mosar/regimes.py}:
\codepath{RegimeKernel.forward}
& Non increasing non GG gates in $[e^{-B},1]$; exact $GG=1$.\newline \\

$W_{bij}=\pi_Q^\top\mathbf G\pi_K$
& \codepath{mosar/worth_field.py}: \codepath{WorthField.forward}
& Bilinear \codepath{einsum}; no pairwise QK signal enters the routers.\newline \\

$B_{bij}=\log\max(W_{bij},\varepsilon)$
&
\codepath{mosar/worth_field.py}:
\codepath{WorthField.forward}
& Clamp, not additive epsilon; pure GG bias must be exactly zero.\newline \\

$S_{ij}^{h}$
& \codepath{gemma2/core_attention.py}:
\codepath{MoSARGemma2DotProduct}
\newline
\codepath{Attention.forward}\newline
& Native Gemma 2 QK scaling is preserved; GQA expansion occurs after routing.\newline \\

$A=S+B+M$
& \codepath{gemma2/core_attention.py}:
\codepath{MoSARGemma2DotProduct}\newline
\codepath{Attention.forward}\newline
& Worth bias is broadcast over heads and added before masking and softmax.\newline \\

$\mathcal L_{\mathrm{cost}}$
& \codepath{mosar/losses.py}: \codepath{normalized_reach_cost}
& Costs equal $\delta_m/N$; Q and K expected costs are averaged with factor $1/2$.\newline \\

$\mathcal L_{\mathrm{LM}}+\lambda\mathcal L_{\mathrm{cost}}$
& \codepath{gemma2/mosar_forward_step.py}: \codepath{MoSARObjectiveState.cost_weight}\newline
& The configured trajectory controls $\lambda$; the cost run uses linear warmup.\newline \\

Hard top 1 routing
& \codepath{gemma2/core_attention.py}: \codepath{MoSARGemma2DotProduct}
\newline
\codepath{Attention._route}\newline
& Q and K router probabilities are replaced by one hot top 1 assignments at inference.\newline \\

$\Omega_\theta$
& \codepath{mosar/hard_domain.py}: \codepath{build_hard_domain}
& Top-1 Q and K regimes define the explicit causal hard domain.\newline \\

Structured outputs
& \codepath{mosar/outputs.py} and reference output dataclasses
& Preserve probabilities, logits, worth, bias, and diagnostics for tests.\newline \\

Reference attention
& \codepath{mosar/reference_attention.py}: \codepath{MoSARReferenceAttention}
& Provides a readable model agnostic implementation for correctness tests.\newline \\

Experimental adapter
& \codepath{integrations/megatron/gemma2/}
& Replaces the Gemma 2 core attention module while preserving the remaining backbone.\newline \\

\bottomrule
\end{longtable}
\endgroup

\section{Compatibility checklist}

A repository revision is mathematically compatible with this specification
only if all checked statements below remain true.

\begin{longtable}{p{0.7cm}p{12.5cm}}
\toprule
 & Requirement \\
\midrule
\endfirsthead
\toprule
 & Requirement \\
\midrule
\endhead
$\square$ & Query routing is computed from post-positional query representations. \\
$\square$ & Key routing is computed from post-positional, unexpanded key representations. \\
$\square$ & Query and key routers are separately parameterized. \\
$\square$ & Each router ends in a temperature-scaled softmax over regimes. \\
$\square$ & Router MLP biases do not bypass the final softmax. \\
$\square$ & The factorized pair weights are $\pi_Q(m)\pi_K(n)$. \\
$\square$ & The regime matrix depends only on relative distance and frozen regime geometry. \\
$\square$ & The positive worth gate is $W=\pi_Q^\top\mathbf G(d)\pi_K$. \\
$\square$ & The additive attention bias is $\log(\max\{W,\varepsilon\})$. \\
$\square$ & The implementation is a mixture of gates, not a direct mixture of additive biases. \\
$\square$ & The worth field is not incorrectly normalized over key positions. \\
$\square$ & The final attention softmax normalizes over valid key positions. \\
$\square$ & The worth bias is shared across heads; QK similarities remain head-specific. \\
$\square$ & Pure GG routing yields exactly zero additional bias. \\
$\square$ & The causal mask remains authoritative. \\
$\square$ & The cost is the role-averaged expected normalized reach. \\
$\square$ & The hard domain is constructible before pairwise QK computation. \\
$\square$ & Backbone-specific and distributed changes are isolated from the mathematical core. \\
\bottomrule
\end{longtable}

\section{Known implementation boundaries}

\subsection{Model agnosticism}

The mathematical operator is not tied to Gemma. It requires an
adapter that can expose the query and key representations used by attention
after the model's token-wise positional transformation, and can add a shared
bias to attention logits before softmax.

For architectures with pairwise positional biases rather than a token-wise
transform such as RoPE, the adapter must explicitly define what representation
the routers observe. Such a choice is an adapter-level methodological decision,
not silently covered by the current post-positional contract.

\subsection{Tensor parallelism}

The current router concatenates locally available heads. Under tensor
parallelism, heads may be partitioned across ranks. A compatible production
implementation must compute the same logical projection as the full
concatenation, for example through a row-parallel linear projection followed by
an all-reduce of partial contributions. It must not make independent,
rank-specific routing decisions for a worth field that is intended to be
shared across all attention heads.

\subsection{Soft decay versus hard sparsity}

The soft training mechanism uses strictly positive gates and therefore induces
continuous decay rather than exact zeros. Exact computational sparsity belongs
to the hard policy and a corresponding sparse kernel. Dense soft training alone
does not establish wall-clock speedup.

\subsection{Reference versus production attention}

The dense reference path repeats KV heads and materializes $O(N^2)$ tensors for
clarity. A production adapter may use fused or grouped operations, provided it
is numerically compatible with the same equations and passes the no-op and
forced-global equivalence tests.

\subsection{Hardware and experimental constraints}
\label{hardware}
All experiments were conducted on a GPU-accelerated Booster node equipped with 64GB NVIDIA A100 accelerators. In our experimental setup, however, each training job was assigned a single GPU, and the longest training trajectories were executed as a sequence of independent SLURM jobs on fixed segments of 50,000 steps. Checkpoints were saved every 10,000 optimization steps, allowing segments interrupted or delayed by the last valid checkpoint to be resumed.

This setup provides a realistic academic large-scale training environment, but it also imposes several practical constraints. First, the experiments are subject to queueing delays, node reservations, scheduler availability, and wall-clock time limits. As a result, the main training trajectories progress asynchronously across variants: lighter baselines often complete a segment earlier, while MoSAR variants require longer wall-clock time because of the additional routing and diagnostic computation. Second, because the cluster is shared and not dedicated to a single project, the number of variants, seeds, intermediate evaluations, and downstream ablations that can be run is constrained by allocation availability rather than by the conceptual scope of the method. Third, the use of segmented training introduces additional engineering overhead, including checkpoint validation, log inspection, recovery from I/O or permission-related failures, and repeated evaluation-only jobs at milestone checkpoints.

These constraints primarily affect the breadth and cadence of the experimental campaign, not the nature of the controlled comparisons. All main variants are trained from the same initialization seed, data seed, tokenizer, packed-token stream, model scale, sequence length, optimization schedule, and evaluation protocol. Therefore, the reported comparisons are intended to isolate the effect of the attention geometry rather than differences in training setup. With larger or dedicated compute, the natural extension would be to increase the number of seeds, evaluate more cost coefficients, run longer extrapolation and hard-inference ablations, and report denser milestone curves. We do not expect these additional experiments to change the central methodological contribution of MoSAR: the proposal of an adaptive, learnable attention geometry. Rather, additional compute would mainly increase the number and granularity of empirical stress tests around the same core mechanism.

\end{document}

%% file: math_commands.tex
\usepackage{amsmath,amsfonts,bm}

\def\eqref#1{equation~\ref{#1}}
\def\1{\bm{1}}

\DeclareMathAlphabet{\mathsfit}{\encodingdefault}{\sfdefault}{m}{sl}
\SetMathAlphabet{\mathsfit}{bold}{\encodingdefault}{\sfdefault}{bx}{n}

\newcommand{\R}{\mathbb{R}}

\newcommand{\softmax}{\mathrm{softmax}}

%% file: figures/training_curve_seq2048_ppl_pgfplots.tex
\begin{tikzpicture}
\begin{axis}[
    width=\linewidth,
    height=0.50\linewidth,
    xlabel={Training steps (k)},
    ylabel={Validation perplexity},
    xmin=50, xmax=610,
    ymin=11.764017, ymax=21.150423,
    grid=both,
    major grid style={line width=.2pt,draw=gray!25},
    minor grid style={line width=.1pt,draw=gray!10},
    legend style={font=\scriptsize, draw=none, fill=none, at={(0.02,0.98)}, anchor=north west},
    tick label style={font=\scriptsize},
    label style={font=\small},
    legend columns=2,
    mark size=1.4pt,
]
\addplot+[color=green!50!black, dashed, mark=pentagon*] coordinates {(50,20.185660) (100,16.602030) (150,16.504230) (200,16.057690) (250,14.507440) (300,14.985050) (350,14.455740) (400,14.275980) (450,13.570900) (500,12.792380) (550,13.058400) (610,12.266860)};
\addlegendentry{Fixed-M}
\addplot+[color=red, very thick, solid, mark=*] coordinates {(50,20.171370) (100,16.643840) (150,16.494020) (200,16.058560) (250,14.496010) (300,14.964440) (350,14.506440) (400,14.261570) (450,13.569530) (500,12.816060) (550,13.086610) (610,12.268730)};
\addlegendentry{MoSAR}
\addplot+[color=orange, dashed, mark=diamond*] coordinates {(50,20.203330) (100,16.728850) (150,16.487000) (200,16.099590) (250,14.544000) (300,15.095670) (350,14.560470) (400,14.310290) (450,13.638840) (500,12.802060) (550,13.107640) (610,12.313070)};
\addlegendentry{Fixed-S}
\addplot+[color=magenta, dash dot, mark=triangle*] coordinates {(50,20.149390) (100,16.707770) (150,16.569450) (200,16.248760) (250,14.681700) (300,15.118050) (350,14.599160) (400,14.387060) (450,13.707410) (500,12.931170) (550,13.238200) (610,12.409410)};
\addlegendentry{0.75-RoPE}
\addplot+[color=blue, solid, mark=square*] coordinates {(50,20.465450) (100,16.934460) (150,16.672780) (200,16.287830) (250,14.704880) (300,15.155120) (350,14.653410) (400,14.428200) (450,13.761700) (500,12.894760) (550,13.173900) (610,12.410710)};
\addlegendentry{ALiBi}
\addplot+[color=black, solid, mark=*] coordinates {(50,20.353810) (100,16.723520) (150,16.761810) (200,16.299180) (250,14.641920) (300,15.237690) (350,14.705180) (400,14.488850) (450,13.787140) (500,12.988030) (550,13.278340) (610,12.517210)};
\addlegendentry{RoPE}
\addplot+[color=cyan!60!black, very thick, dotted, mark=otimes*] coordinates {(50,20.630880) (100,16.892500) (150,16.836530) (200,16.355720) (250,14.850140) (300,15.317450) (350,14.744570) (400,14.596700) (450,13.890960) (500,13.073890) (550,13.357040) (610,12.552620)};
\addlegendentry{MoSAR-Cost}
\addplot+[color=gray, dash dot, mark=x] coordinates {(50,20.647580) (100,16.777200) (150,17.044940) (200,16.510170) (250,15.011420) (300,15.471250) (350,14.967000) (400,14.686470) (450,14.008880) (500,13.200140) (550,13.495460) (610,12.703480)};
\addlegendentry{RoPE-M-mask}
\end{axis}
\end{tikzpicture}